\documentclass[11pt]{article}
\usepackage{algorithm}
\usepackage{algorithmic}
\usepackage{amsmath}
\usepackage{amssymb}
\usepackage{authblk}
\usepackage{booktabs}
\usepackage[margin=1in]{geometry}
\usepackage{graphicx}
\usepackage{mathtools}
\usepackage{ulem}
\usepackage[colorlinks=true,citecolor=blue]{hyperref}

\title{RED-DiffEq: Regularization by denoising diffusion models for solving inverse PDE problems with application to full waveform inversion}

\author[1]{Siming Shan}
\author[1]{Min Zhu}
\author[2]{Youzuo Lin}
\author[1,*]{Lu Lu}

\affil[1]{Department of Statistics and Data Science, Yale University, New Haven, CT 06511, USA}
\affil[2]{School of Data Science and Society, University of North Carolina at Chapel Hill, Chapel Hill, NC 27599, USA}
\affil[*]{Corresponding author. Email: lu.lu@yale.edu}
\date{}

\begin{document}
%\linenumbers
\maketitle

\begin{abstract}
Partial differential equation (PDE)-governed inverse problems are fundamental across various scientific and engineering applications; yet they face significant challenges due to nonlinearity, ill-posedness, and sensitivity to noise. Here, we introduce a new computational framework, regularization by denoising using diffusion models for partial differential equations (RED-DiffEq), by integrating physics-driven inversion and data-driven learning. RED-DiffEq leverages pretrained diffusion models as a regularization mechanism for PDE-governed inverse problems. We apply RED-DiffEq to solve the full waveform inversion problem in geophysics, a challenging seismic imaging technique that seeks to reconstruct high-resolution subsurface velocity models from seismic measurement data. Our method shows enhanced accuracy and robustness compared to benchmark methods. Additionally, it exhibits strong generalization and domain decomposition capacity, enabling the inversion of more complex velocity models with larger domains than those used in training the diffusion model. Our framework can also be directly applied to diverse PDE-governed inverse problems.
\end{abstract}

\section{Introduction}
\label{sec:intro}

Inverse problems governed by partial differential equations (PDEs) play a crucial role in diverse scientific and engineering domains~\cite{inverse_problem,tarantola2005,ghattas2019,lu2020extraction}, from medical imaging~\cite{ammari2008} and fluid dynamics~\cite{Cai2022PhysicsInformedNN} to geophysical studies~\cite{Sambridge2002MonteCM}. These problems involve inferring unknown parameters or fields from indirect observations, where the forward process is governed by PDEs. Unlike linear inverse problems, PDE-governed inverse problems usually approximate the solution as a nonconvex and nonlinear optimization task iteratively minimizing the mismatch between observed data and the current PDE solution to optimize unknown parameters. A prominent example is full waveform inversion (FWI) in seismic imaging, which aims to infer subsurface structures for resource exploration, environmental studies, and seismic hazard assessment~\cite{fwi_review, Zhang2021Rayleigh}. In FWI, seismic waves are generated at the Earth’s surface and recorded after propagating through the subsurface. The goal is to reconstruct spatially varying wave-speed (velocity) fields by matching simulated waveforms to recorded measurements.

Such inverse problems are intrinsically challenging due to their nonlinearity and ill-posedness, rendering them highly sensitive to data imperfections like measurement noise and incomplete observations~\cite{ZhangFWI2020,yazdani2020systems,daneker2023systems,jiao2021one}. In FWI, a distinct and critical challenge is cycle skipping, which arises when the predicted and observed waveforms are misaligned by more than half a period. This misalignment causes the local optimization to match incorrect phases, trapping the solution in a spurious basin of attraction~\cite{pladys2021cycle}. Consequently, cycle skipping is the primary mechanism driving convergence to incorrect local minima, while measurement noise and missing data further exacerbate the instability and ill-posedness of the inversion. To mitigate ill-posedness and enhance robustness against imperfect observations (e.g., noise, limited data), various regularization strategies have been employed. Classical approaches, such as Tikhonov regularization, enforce smoothness constraints on the velocity models~\cite{piror}, whereas Total Variation (TV) regularization promotes piecewise-constant structures to preserve sharp interfaces~\cite{total_variation, tv_lin}. These regularizers stabilize the inversion by penalizing geologically implausible high-frequency artifacts and constraining the null space, often resulting in more robust optimization behavior. As such, Tikhonov and TV regularization serve as standard baselines in full waveform inversion~\cite{tv_l2}.

Machine learning has introduced innovative approaches to PDE-governed inverse problems. Deep learning models, particularly convolutional neural networks, have demonstrated success in learning direct mappings from observations to underlying parameters in synthetic environments~\cite{youzuo_study, inversionnet, velocitygan, Kazei2021Mapping}. These approaches offer potential advantages in computational efficiency and eliminate sensitivity to initial parameters~\cite{Doe2024}. However, their practical application is often limited by poor generalization, especially when handling noise, missing data, or scenarios not represented in training datasets~\cite{Martinsson2021Neural, Karnakov2023Solving, Zhang2023}.
To address these limitations, advanced physics-informed machine learning frameworks have emerged. Physics-informed neural networks (PINNs) integrate physical constraints directly into the learning process, demonstrating promising results for solving inverse problems~\cite{karniadakis2021physics, lu2021physics,lu2021deepxde}. This integration improves the consistency of the solution with governing physical laws while preserving computational efficiency~\cite{chen2020physics,zhang2019quantifying,pang2019fpinns,Catricheo2024}. In the geophysical domain, PINNs have been actively explored for both wave-equation modeling and inversion, demonstrating their capacity to embed wave propagation physics directly into data-driven frameworks~\cite{song_alphalfiah_2021_gji,huang_alkhalifah_2022_pinnup,huang_alkhalifah_2021_positional,huang_alkhalifah_2023_hashpinn,song_alkhalifah_2021_wri}.
Building on this foundation, deep neural operators~\cite{deeponet} have advanced the field by learning sophisticated mappings between function spaces, offering improved generalization across diverse physical scenarios~\cite{deepmionet, geocarbon_deeponet,di2023neural,mao2024disk2planet,jiao2024solving,kou2025neural}. Neural operators have also been studied for full waveform inversion~\cite{fourierdeeponet,ma_wang_huang_alkhalifah_2025_pi_fwiner}. Despite these significant advances, both approaches continue to face challenges with out-of-distribution data, particularly when confronted with complex real-world problems.

Another promising recent development in machine learning has been the emergence of diffusion models, which have revolutionized generative modeling through their iterative denoising approach to data generation, effectively capturing complex probability distributions~\cite{scorebased, ddpm, ddim}. These models offer unique advantages for inverse problems due to their ability to incorporate prior information and their inherent robustness to noise~\cite{ddrm, song2023pseudoinverseguided, dou2024diffusion,wang2025fundiff}. Diffusion models have also demonstrated success in reconstructing complex fields from limited or noisy observations in various domains. Compared to PINNs and neural operators, diffusion-model-based solvers are inherently less sensitive to out-of-distribution scenarios, and more robust against noise and sparse data. More importantly, they eliminate the need for requiring paired data for training and offer a more flexible framework for incorporating numerical solvers during the sampling process to regularize the result~\cite{pird, physics_informed_diffusion, red-diff}. However, their application to inverse problems involving explicit PDE solvers as forward operators remains under-explored. In FWI, initial efforts have primarily focused on embedding the conventional FWI update step within the sampling process of a diffusion model. For instance, Wang \textit{et al.} proposed DiffusionFWI that integrates FWI iterations into the generative sampling trajectory to solve acoustic FWI~\cite{Wang_2023}. Taufik \textit{et al.} extended this to elastic FWI~\cite{diffusionefwi} and subsequently introduced Iterative Latent Variable Refinement (ILVR) to enhance reconstruction quality~\cite{ILVR, diffusionilvr}. Other studies have investigated joint diffusion architectures that solve for velocity models directly in a latent space~\cite{wang2024}. While these approaches establish the viability of diffusion models for FWI, they predominantly embed conventional FWI updates within the diffusion sampling trajectory. This formulation may make it difficult to systematically balance data fidelity against the learned prior.

These observations motivate the need for a framework that combines physics-based inversion with strong data-driven priors while retaining scalability and generalization. Here, we develop a general framework, regularization by denoising using diffusion models for partial differential equations (RED-DiffEq), that integrates diffusion models directly into PDE-governed inverse problems as a regularization mechanism. Our approach leverages diffusion models to learn robust prior distributions over plausible solutions from synthetic datasets. By employing the pretrained diffusion model as a physics-aware regularization term, the regularization effect is achieved by calculating the residual between actual and predicted noise. Moreover, RED-DiffEq incorporates a flexible and scalable domain decomposition capability. By utilizing the Domain Decomposed Regularization (DDR) strategy, the framework allows for training on small sub-domains and seamless deployment on substantially larger domains without retraining.

We demonstrate the effectiveness of RED-DiffEq through extensive validation on FWI, a challenging inverse problem in geophysics. Our results consistently show superior performance in accuracy and robustness compared to conventional regularization methods and existing diffusion-based approaches, both quantitatively and qualitatively. Notably, when trained on the OpenFWI dataset (a large-scale synthetic seismic benchmark providing paired velocity models and simulated wavefield recordings~\cite{openfwi}), RED-DiffEq demonstrates strong generalization capabilities to benchmark models such as Marmousi~\cite{marmousi} and Overthrust~\cite{overthrust} enabled by domain decomposition. This highlights its potential for application to real-world field data.

\section{Results}
\label{sec:results}

We first introduce full waveform inversion in Section~\ref{sec:fwi_intro} and present the RED-DiffEq framework in Section~\ref{sec:red-diffeq_fwi}. We then evaluate the effectiveness of RED-DiffEq in four scenarios from the OpenFWI benchmark (Section~\ref{sec:openfwi_result}). We also analyze the uncertainty quantification capabilities of our method (Section~\ref{sec:uq_result}). Finally, we test RED-DiffEq on the Marmousi and Overthrust models (Section~\ref{sec:marmousi_result}) to demonstrate its generalizability and domain decomposition capabilities. All the experimental details, including the training of the diffusion model, hyperparameter tuning, and other design choices are presented in Section~\ref{sec:appendix_experiment_detail}.

\subsection{Full waveform inversion}
\label{sec:fwi_intro}

\begin{figure}[htbp]
    \centering
    \includegraphics[width=1\linewidth]{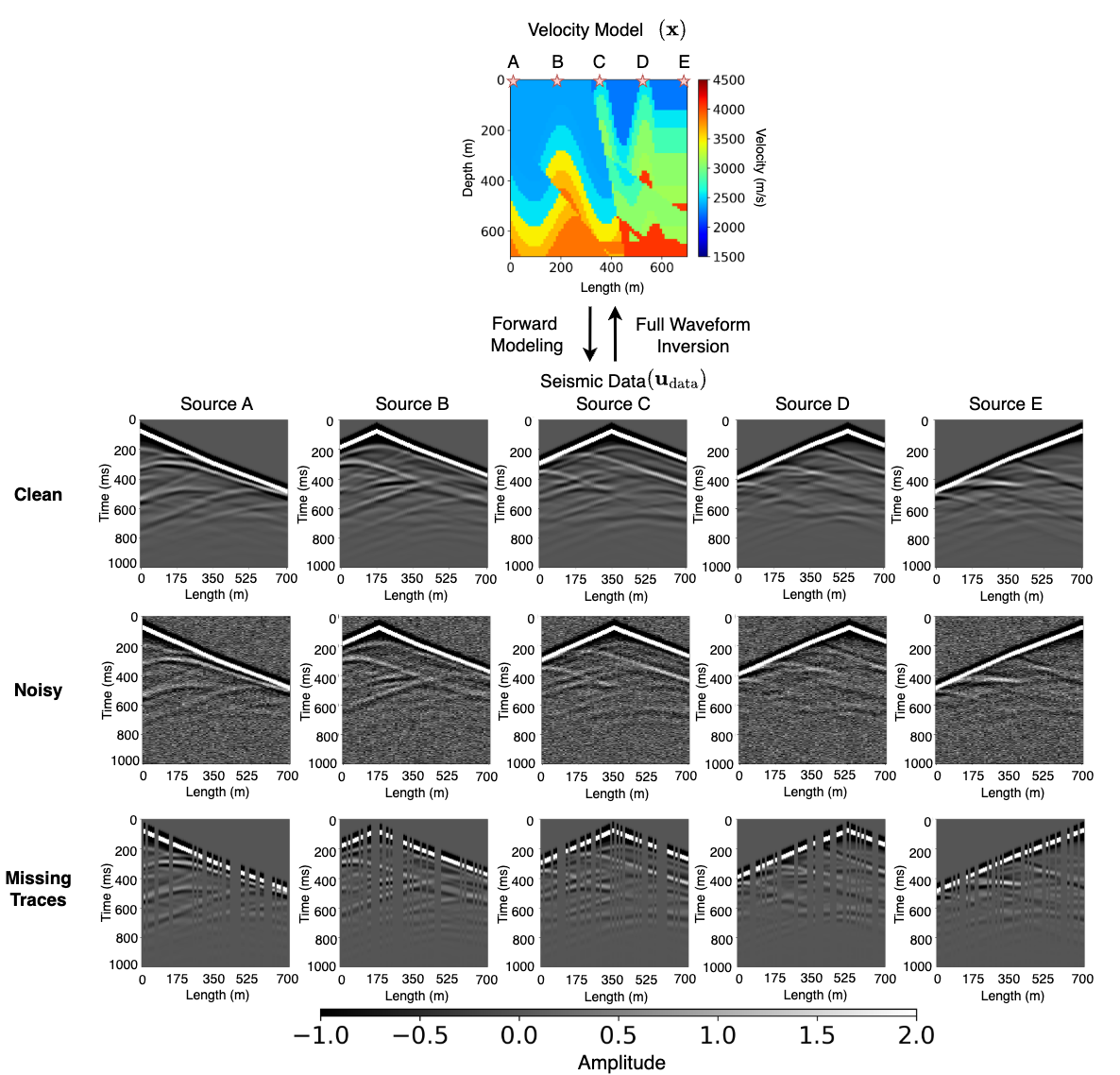}  
    \caption{\textbf{Forward modeling and full waveform inversion.} The velocity model (top) with five sources (A--E) generates seismic data under different conditions: clean data (noise-free), noisy data, and data with missing traces.}
    \label{fig:openfwi}
\end{figure}

Full waveform inversion seeks to reconstruct high-resolution subsurface velocity models from wavefield measurement data (potentially incomplete or noisy) by leveraging the physics of wave propagation (Fig.~\ref{fig:openfwi}). In the acoustic approximation, wave propagation is governed by
\begin{equation}
\label{eq:wave_eq}
\frac{1}{\mathbf{x}^2(\mathbf{r})} \frac{\partial^2 \mathbf{u}(\mathbf{r},t)}{\partial t^2} \;-\; \nabla^2 \mathbf{u}(\mathbf{r},t) \;=\; q(\mathbf{r},t),
\end{equation}
where $\mathbf{r}$ represents the spatial coordinate, $t$ is time, $\nabla^2$ is the spatial Laplacian, and \(\mathbf{x}(\mathbf{r})\) denotes the subsurface velocity model, \(\mathbf{u}(\mathbf{r},t)\) is the seismic wavefield, and \(q(\mathbf{r},t)\) represents the source term. The process of solving this equation for a given velocity model to predict the wavefield is known as forward modeling.

Starting from an initial velocity model $\mathbf{x}$ (typically a smoothed or rough approximation of the true geology), traditional physics-driven numerical methods for FWI are formulated as an optimization problem that iteratively adjusts the velocity model $\textbf{x}$ to minimize an objective function, which reconciles simulated data with observed data:
\begin{equation}
\label{eq:fwi_objective}
\underset{\mathbf{x}}{\text{argmin}} \ \| \mathbf{u}_{\text{data}} - f_{\text{PDE}}(\mathbf{x}) \|^2_2 + \lambda R(\mathbf{x}).
\end{equation}
Here, the first term measures the misfit between the observed seismic data, $\mathbf{u}_{\text{data}}$, and the simulated data, $f_{\text{PDE}}(\mathbf{x})$, obtained from solving Eq.~\eqref{eq:wave_eq}. The second term is a regularization term weighted by $\lambda$, which is crucial for constraining the solution to be geologically plausible and for improving the convergence of the inversion.

While traditional regularization methods, such as Tikhonov and TV, improve performance, they often fail to produce geologically realistic models. Tikhonov regularization, for instance, tends to introduce over-smoothing, while TV can create ``staircase'' artifacts. Moreover, this physics-driven approach leads to an ill-posed inverse problem that is highly sensitive to the choice of regularization method and to the presence of measurement noise that is typically present in seismic measurement data (e.g., Gaussian noise modeling thermal fluctuations or Laplacian noise modeling impulsive outliers with heavy-tailed statistics). For a more detailed description of FWI, please refer to Section~\ref{sec:appendix_FWI}.

\subsection{RED-DiffEq for full waveform inversion}
\label{sec:red-diffeq_fwi}
To solve inverse problems of PDEs like FWI, we develop a new regularization framework named RED-DiffEq (Algorithm~\ref{alg:fwi-diffq}) based on the REgularization by Denoising (RED) method~\cite{RED2017}. Let $p_{\text{data}}$ denote the probability density of the prior distribution over clean, geologically plausible velocity models. For a velocity model $\mathbf{x}$ at iteration $k$ (denoted $\mathbf{x}_k$), the RED regularizer is
\begin{equation}
\label{eq:red_regularizer}
R(\mathbf{x}_k) \;=\; \mathbb{E}_{t,\boldsymbol{\epsilon}}[\mathbf{x}_k^\top\!\bigl(\mathbf{x}_k-\mathcal{D}_\theta(\mathbf{x}_k;t,\boldsymbol{\epsilon})\bigr)].
\end{equation}
$\mathcal{D}_\theta$ is a Tweedie-inspired denoising operator that nudges $\mathbf{x}_k$ toward regions of higher density.
To define $\mathcal{D}_\theta$, let $\gamma(t)\!\in\!(0,1]$ be a noise schedule (Algorithm~\ref{alg:sigmoid_noise_schedule}), where $t\sim\mathcal{U}\{1{:}T\}$, and $T$ is a hyperparameter. In each iteration of RED-DiffEq, we first compute a noisy velocity model $\mathbf{x}_{k,t}$ from the Variance-Preserving (VP) corruption function by
\begin{equation}
\label{eq:vp_noise}
\mathbf{x}_{k,t} \;=\; \sqrt{\gamma(t)}\,\mathbf{x}_k \;+\; \sqrt{1-\gamma(t)}\,\boldsymbol{\epsilon},
\quad
\boldsymbol{\epsilon}\!\sim\!\mathcal{N}(\mathbf{0},\mathbf{I}).
\end{equation}
Then, by Tweedie’s identity for Gaussian channels, we set
\begin{equation}
\label{eq:tweedie_identity}
\mathcal{D}_\theta(\mathbf{x}_{k};t,\boldsymbol{\epsilon})
\;=\;
\frac{1}{\sqrt{\gamma(t)}}\Bigl(\mathbf{x}_{k,t} \;+\; \bigl(1-\gamma(t)\bigr)\, s_{\text{data},t}(\mathbf{x}_{k,t})\Bigr).
\end{equation}
In particular, $s_{\text{data},t}(\mathbf{z}) = \nabla_{\mathbf{z}}\log p_{\text{data},t}(\mathbf{z})$, where $p_{\text{data},t}$ is the noisy marginal density obtained by applying the VP corruption in Eq.~\eqref{eq:vp_noise} to samples from the distribution with the density $p_{\text{data}}$.

\begin{figure}[htbp]
    \centering
    \includegraphics[width=1\linewidth]{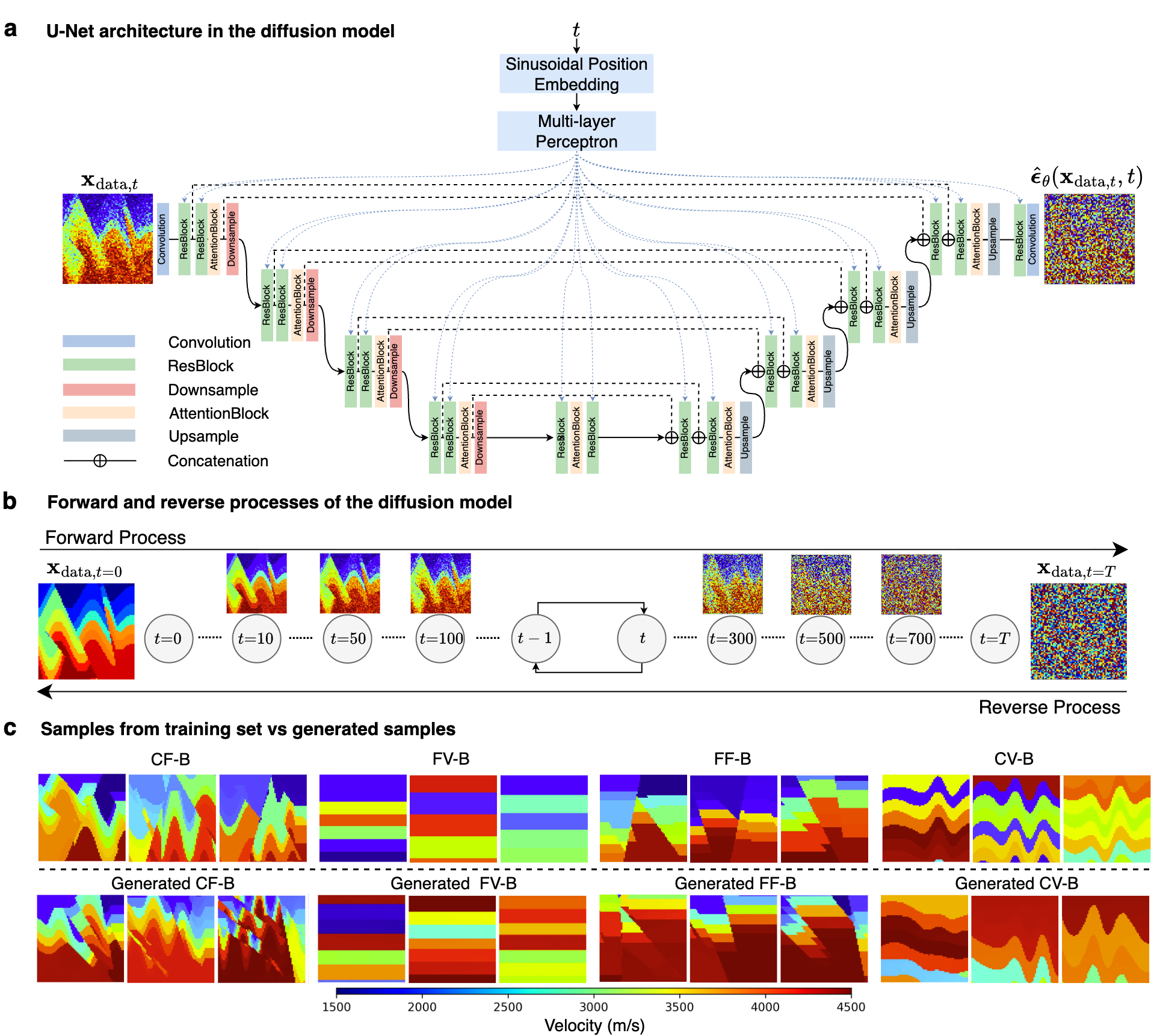}  
    \caption{\textbf{Diffusion model architecture and generated samples of velocity map.} (\textbf{a}) Schematic of the U-Net denoising network used in the diffusion model. A noisy velocity model is processed through an encoder–decoder U-Net with ResBlocks, downsampling and upsampling layers. Sinusoidal time embeddings are passed through a multilayer perceptron and injected into each ResBlock to condition the network on the diffusion timestep. The network predicts the added noise. (\textbf{b}) Overview of the complete diffusion process, showing both the forward noising process and the learned reverse denoising process. In this study, we set the maximum diffusion time step $T = 1000$ (\textbf{c}) Comparison between velocity maps from the training dataset (top row) and unconditionally generated velocity maps from a single diffusion model pretrained with four velocity model families altogether (bottom row).}
    \label{fig:diffusion}
\end{figure}

\begin{figure}[htbp]
    \centering
    \includegraphics[width=1\linewidth]{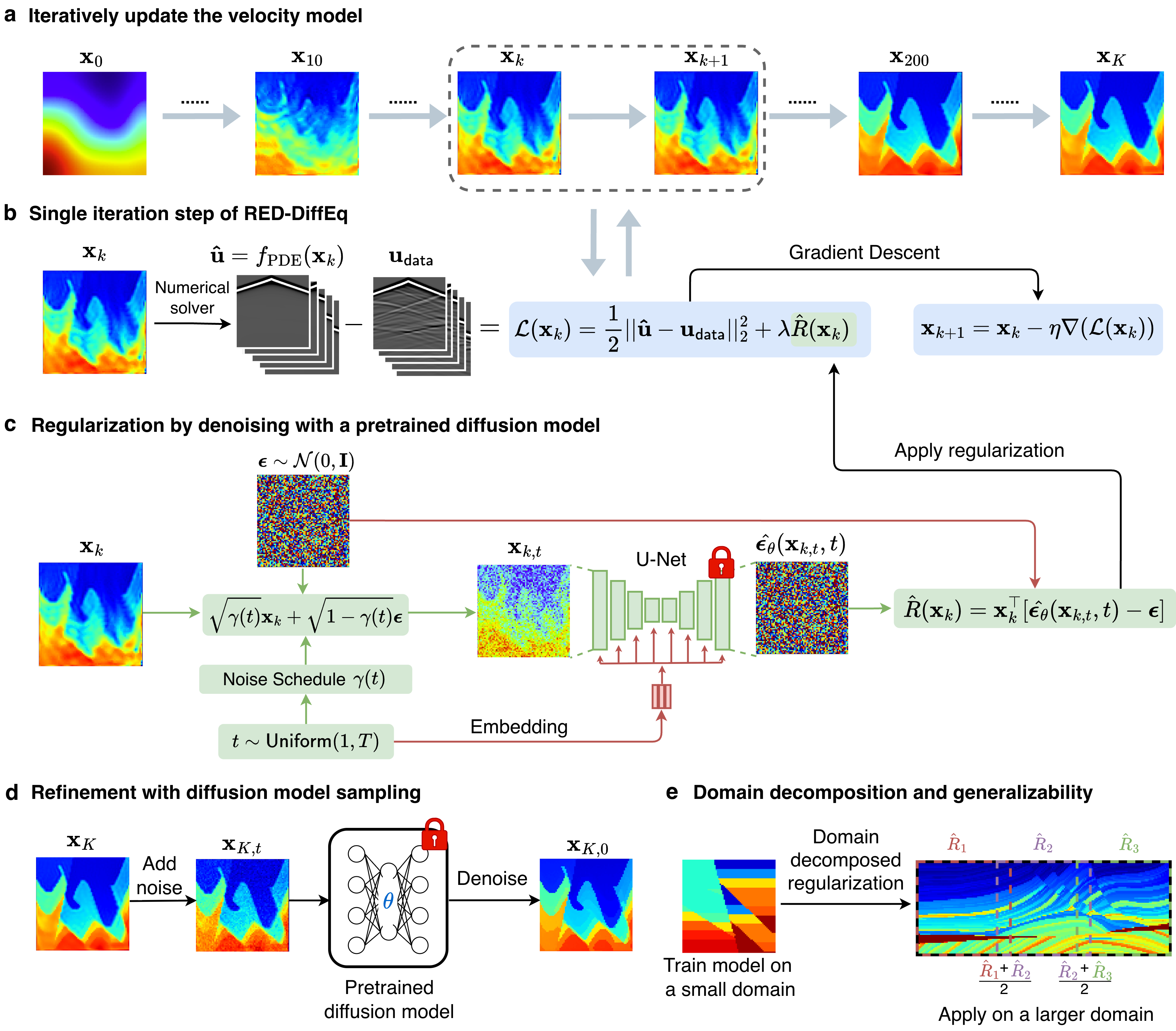}  
    \caption{
    \textbf{Schematic illustration of RED-DiffEq for full waveform inversion.} (\textbf{a}) An overview of the inversion process that iteratively updates the velocity model. (\textbf{b}) An illustration of each iteration step of RED-DiffEq. (\textbf{c}) Calculation of the diffusion-based regularization term. (\textbf{d}) Optional post-processing refinement step using the pretrained diffusion model to further refine the velocity map after the main inversion process is finished. (\textbf{e}) Domain Decomposed Regularization (DDR) strategy for RED-DiffEq. We can train the diffusion model on a small domain and apply on a larger domain during inversion using sliding windows to compute the regularization.
    }
    \label{fig:diff_fwi_framework}
\end{figure}

We approximate this score function by a neural network $\hat{\boldsymbol{\epsilon}}_{\theta}$ (typically a U-Net as in Fig.~\ref{fig:diffusion}a) using the Denoising Diffusion Probabilistic Model (DDPM) on a corpus of clean velocity models $\mathbf{x}_{\text{data}}$ (Fig.~\ref{fig:diffusion}b). During training, the network is optimized to predict the specific noise $\boldsymbol{\epsilon}$ added to the input during the VP corruption process. After training, we estimate the score in Eq.~\eqref{eq:tweedie_identity} as
\begin{equation}
\label{eq:epsilon_score}
s_{\text{data},t}(\mathbf{x}_{k,t}) \;\approx\; -\frac{\hat{\boldsymbol{\epsilon}}_\theta(\mathbf{x}_{k,t},t)}{\sqrt{1-\gamma(t)}}.
\end{equation}
Then substituting Eq.~\eqref{eq:epsilon_score} into Eq.~\eqref{eq:tweedie_identity} yields the practical denoise operator
\begin{equation}
\label{eq:learned_denoiser}
\mathcal{D}_\theta(\mathbf{x}_{k};t,\boldsymbol{\epsilon})
\;=\;
\frac{1}{\sqrt{\gamma(t)}}\Bigl(\mathbf{x}_{k,t} \;-\; \sqrt{1-\gamma(t)}\,\hat{\boldsymbol{\epsilon}}_\theta(\mathbf{x}_{k,t},t)\Bigr).
\end{equation}
Finally, substituting Eq.~\eqref{eq:learned_denoiser} into Eq.~\eqref{eq:red_regularizer} gives the RED regularizer as
\begin{equation}
R(\mathbf{x}_k)
\;=\;
\mathbb{E}_{t,\boldsymbol{\epsilon}}
\left[
w(t)\,
\mathbf{x}_{k}^\top\!\Bigl(\hat{\boldsymbol{\epsilon}}_\theta(\mathbf{x}_{k,t},t)-\boldsymbol{\epsilon}\Bigr)
\right],
\quad
w(t)\;=\;\sqrt{\tfrac{1-\gamma(t)}{\gamma(t)}},
\end{equation}
where $\boldsymbol{\epsilon}$ is the random Gaussian noise we draw in Eq.~\eqref{eq:vp_noise}. In our experiments, we adopt a pragmatic variant that drops $w(t)$ and uses a constant $\lambda$, which empirically improves stability and convergence. Therefore, the empirical inversion objective minimized at iteration $k$ is given by
\begin{equation}
\mathcal{L}(\mathbf{x}_k)
\;=\; 
\bigl\|\,\mathbf{u}_{\text{data}} - f_{\mathrm{PDE}}(\mathbf{x}_k)\,\bigr\|_2^2
\;+\;
\lambda\,\underbrace{\mathbf{x}_k^\top\!\Bigl(\hat{\boldsymbol{\epsilon}}_\theta(\mathbf{x}_{k,t},t)-\boldsymbol{\epsilon}\Bigr)}_{\hat{R}(\mathbf{x}_k)},
\end{equation}
where $\hat{R}$ provides a Monte Carlo (MC) estimator of the RED regularizer (Fig.~\ref{fig:diff_fwi_framework}).
In practice, we randomly sample $(t,\boldsymbol{\epsilon})$ at each iteration, which injects stochasticity into the inversion process. We present the details of RED-DiffEq in Sec.~\ref{sec:red-diffeq}.

For large-scale inversion domains, we develop the Domain Decomposed Regularization (DDR) strategy (Fig.~\ref{fig:diff_fwi_framework}e). Specifically, during the inversion process, the large target domain is partitioned into smaller, overlapping subdomains. The diffusion-based regularization is computed locally on each subdomain using the fixed, pretrained diffusion model, and these overlapping regularizations are subsequently aggregated to form a consistent global regularization. This design effectively decouples the training scale from the inversion scale, allowing a model trained on small domains to be seamlessly deployed on arbitrarily large domains without retraining.

After the inversion process, we optionally apply a post-processing refinement step. Specifically, the inverted velocity model is perturbed to a selected diffusion timestep $t$ using the VP corruption function (Eq.~\eqref{eq:vp_noise}), and then denoised unconditionally using the pretrained diffusion model. This refinement produces visually cleaner results, at the cost of a minor degradation in quantitative metrics (Sec.~\ref{sec:post_process}). All the results shown in Sec.~\ref{sec:results} are the results without the post-processing refinement step.

\subsection{Validation on the OpenFWI benchmark}
\label{sec:openfwi_result}

We conducted a comprehensive evaluation of RED-DiffEq on the OpenFWI benchmark~\cite{openfwi} under three distinct scenarios: clean seismic data, seismic data with Gaussian noise contamination, and seismic data with missing traces. The OpenFWI dataset comprises pairs of velocity models and simulated seismic wavefields representing diverse geological structures, providing a comprehensive benchmark for evaluating seismic inversion methods. For our experiments, we selected four particularly challenging families from this dataset: Curve Fault (CF-B), Flat Velocity (FV-B), Flat Fault (FF-B), and Curve Velocity (CV-B) (see some samples in Fig.~\ref{fig:diffusion}c). Each family exhibits distinct geological structures that are representative of complex subsurface formations encountered in real-world scenarios. Specifically, CF-B features intricate curved fault systems with varying angles and intersections. FV-B consists of horizontally layered velocity structures. FF-B contains straight fault lines cutting through the model. Lastly, CV-B exhibits curved velocity variations with smooth transitions between layers. For benchmarking, we compared our method against established physics-driven approaches, including standard FWI (no regularization), FWI with Tikhonov regularization, FWI with total variation (TV) regularization, and recent diffusion-model-based methods for FWI~\cite{Wang_2023, diffusionilvr}.

\subsubsection{Clean seismic data}

\begin{figure}[htbp]
    \centering
    \includegraphics[width=1\linewidth]{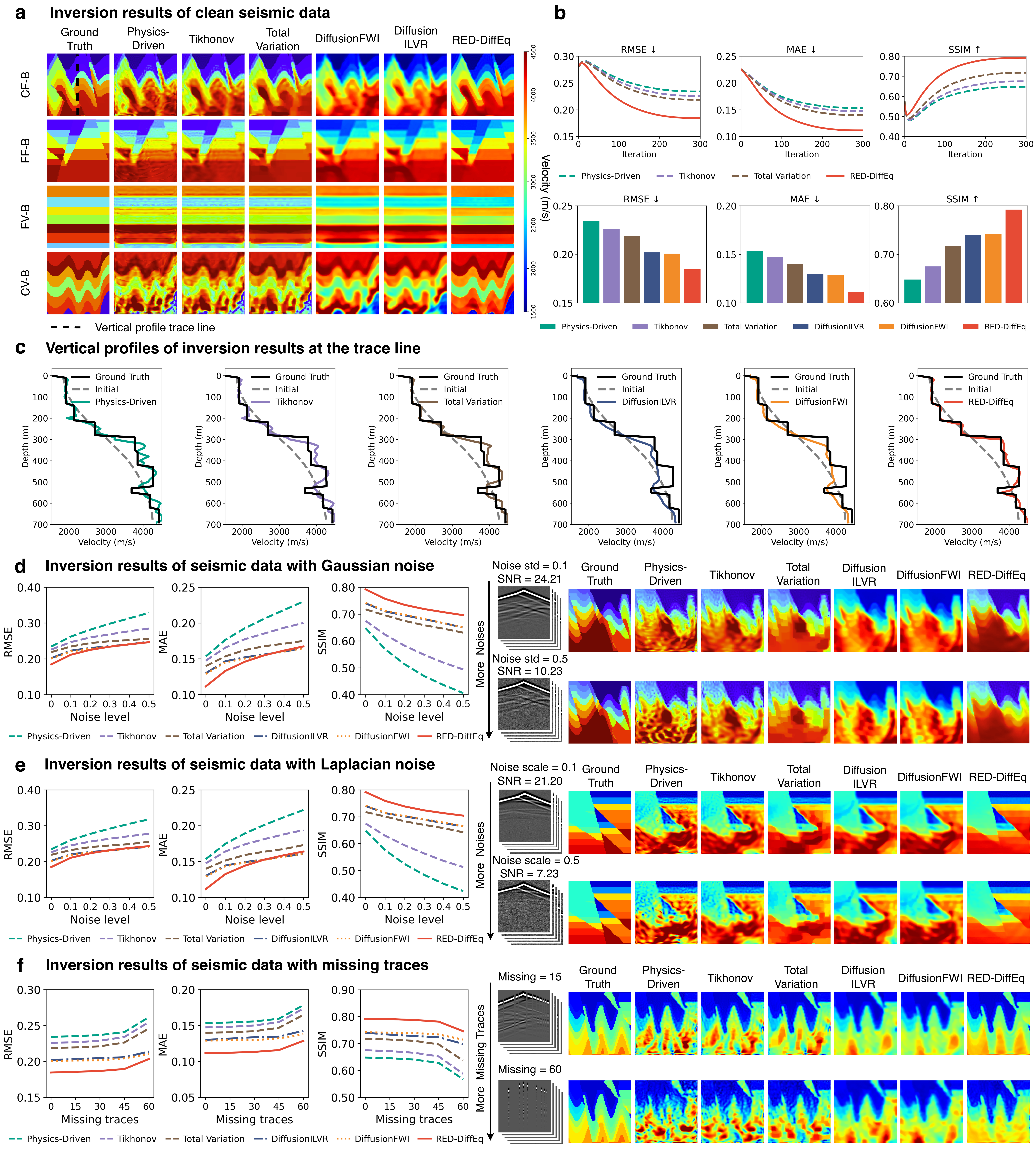}  
    \caption{
    \textbf{Results on the OpenFWI dataset.} (\textbf{a}) Qualitative comparison among different methods. (\textbf{b}) Quantitative comparison among different methods. (\textbf{c}) Vertical profile results of different methods. (\textbf{d}) Performance of different methods under different Gaussian noise levels in the seismic data. (\textbf{e}) Performance of different methods under different Laplacian noise levels in the seismic data. (\textbf{f})  Performance of different methods under different number of missing traces in the seismic data.
    }
    \label{fig:result_openfwi}
\end{figure}

We first evaluated various methods on clean seismic data using 100 previously unseen samples from each of the four chosen geological families in the OpenFWI dataset. All the quantitative metrics are the average of these 400 test cases. For optimization-based approaches (RED-DiffEq, Tikhonov (Sec.~\ref{sec:tikhonov_regularization}), and TV (Sec.~\ref{sec:tv_regularization})), each sample was optimized for 300 iterations. For the diffusion-sampling-based baselines (DiffusionFWI and its variant with the ILVR technique, namely DiffusionILVR), we adopted the protocol suggested in the original implementation~\cite{Wang_2023}, initializing the reverse diffusion process at $t=100$ with 10 conventional FWI gradient steps per diffusion timestep; please refer to Sec.~\ref{sec:hyperparameters_tuning} for more details about hyper-parameters and design choice, and Sec.~\ref{sec:forward_solver} for details about the forward solver used for FWI. All the methods started from an initial model obtained by applying a Gaussian filter (with a standard deviation $\sigma = 10$) to the ground-truth velocity model. Fig.~\ref{fig:result_openfwi}a shows representative velocity inversion examples across the four different geological families. The unregularized standard FWI method produces a large error and fails to capture essential geological features. While Tikhonov regularization improves upon the standard method, it produces an overly smoothed model that loses critical details at layer boundaries. TV preserves sharp edges but introduces ``staircase'' artifacts that distort the geology. DiffusionFWI and DiffusionILVR show promise with simple structures but struggle with fine-scale structures. Our RED‑DiffEq framework outperforms all baselines across the four geological families, simultaneously preserving high‑contrast velocity discontinuities, such as fault planes, while maintaining the smooth lateral and vertical velocity gradients that characterize individual strata. RED‑DiffEq produces the most faithful reconstructions, even for the complex CV‑B models (Fig.~\ref{fig:result_openfwi}a, last row).

The convergence analysis (Fig.~\ref{fig:result_openfwi}b, top row) reveals that RED-DiffEq consistently achieves the lowest root mean squared error (RMSE) and mean absolute error (MAE), and highest  structural similarity index measure (SSIM) throughout the optimization process. We note that DiffusionFWI and DiffusionILVR utilize a nested generation process distinct from standard optimization iterations, rendering a direct convergence curve comparison inapplicable. We compare the final inversion performance across all methods (Fig.~\ref{fig:result_openfwi}b, bottom row), where RED-DiffEq consistently outperforms all benchmarks.

We further evaluated the vertical profiles of the selected inversion examples (Fig.~\ref{fig:result_openfwi}a, top row) by extracting velocity traces at the midpoint (350 m) of the velocity map. The resulting profiles (Fig.~\ref{fig:result_openfwi}c) demonstrate that RED-DiffEq outperforms the other methods by more closely matching the ground truth. Notably, even in depth intervals where all methods perform reasonably well (e.g., from 0 m to 300 m), other methods exhibit oscillations around the ground truth, whereas RED-DiffEq effectively suppresses these artifacts.

\subsubsection{Noisy seismic data}

We evaluated the robustness of various inversion methods by adding Gaussian noise and Laplacian noise with standard deviations and scales ranging from 0.1 to 0.5 to the seismic data. Specifically, Gaussian noise levels (standard deviations ranging from 0.1 to 0.5) correspond to an average Signal-to-Noise Ratio (SNR) range of 24.21 dB to 10.23 dB. Similarly, Laplacian noise levels (scales ranging from 0.1 to 0.5) correspond to an SNR range of 21.20 dB to 7.23 dB. In scientific inversion, an SNR that is around or less than 10 dB is considered challenging in general.
The comparative results reveal distinct behaviors across different approaches under different noisy conditions (Figs.~\ref{fig:result_openfwi}d and e).

All diffusion-model-based methods exhibit noise resilience, and RED-DiffEq more accurately reconstructs distinct layered structures and preserves the continuous interfaces that are evident in the ground truth. In contrast, the unregularized FWI approach shows significant noise-induced artifacts, particularly in the deeper regions of the velocity model, with increasing deterioration at higher noise levels. Tikhonov regularization reduces the impact of measurement noise but introduces excessive smoothing, blurring the sharp boundaries between layers present in the ground truth. While total variation regularization preserves some sharp boundaries, it produces notable staircase artifacts that erase finer details, especially when the noise level reaches 0.5, which further hinders the geological interpretability of the result. 

\subsubsection{Seismic data with missing traces}

We evaluated the performance of all inversion methods in missing-trace scenarios by randomly removing 15 to 60 traces from the seismic data. Since each source location (shot gather) contains 70 traces in total, this corresponds to losing between 21.4\% and 85.7\% of the available acquisition information. Crucially, the specific indices of the missing traces were held constant across all source gathers for a given velocity model. That is, the set of active receiver locations remained identical for every shot in the experiment.

As the proportion of missing data increases, all methods experience performance degradation (Fig.~\ref{fig:result_openfwi}f). However, diffusion-model-based methods exhibit significantly stronger robustness to sparse acquisition compared to standard baselines. RED-DiffEq consistently achieves the lowest RMSE and MAE and the highest SSIM across all missing-trace ratios, demonstrating superior stability under incomplete seismic measurements.

Qualitative comparisons in Fig.~\ref{fig:result_openfwi}f further confirm these observations. When a moderate number of traces are missing (15 traces), RED-DiffEq successfully reconstructs layered structures and continuous interfaces that closely resemble the ground truth. Even in the extreme case of 60 missing traces, RED-DiffEq preserves coherent subsurface structures and suppresses severe artifacts, whereas most benchmark methods fail to recover meaningful geological features. In contrast, the unregularized FWI approach suffers from strong instability, while Tikhonov and Total Variation regularization methods produce overly smooth or staircase-like reconstructions that obscure fine-scale details. These results demonstrate that RED-DiffEq effectively mitigates the ill-posedness introduced by sparse acquisition, providing reliable reconstructions even under severely incomplete seismic data.

\subsection{Uncertainty quantification of RED-DiffEq}
\label{sec:uq_result}

We evaluate the uncertainty quantification (UQ) capability of RED-DiffEq on the OpenFWI benchmark. For each seismic sample, we generated an ensemble of 20 reconstructions by executing the RED-DiffEq inversion with different random seeds, explicitly leveraging the stochastic sampling inherent in the diffusion-based regularizer. We adopt the ensemble mean as the final predicted velocity model and utilize the pixel-wise standard deviation as the metric for spatial uncertainty.

\begin{figure}[t!]
\centering
\includegraphics[width=1.0\linewidth]{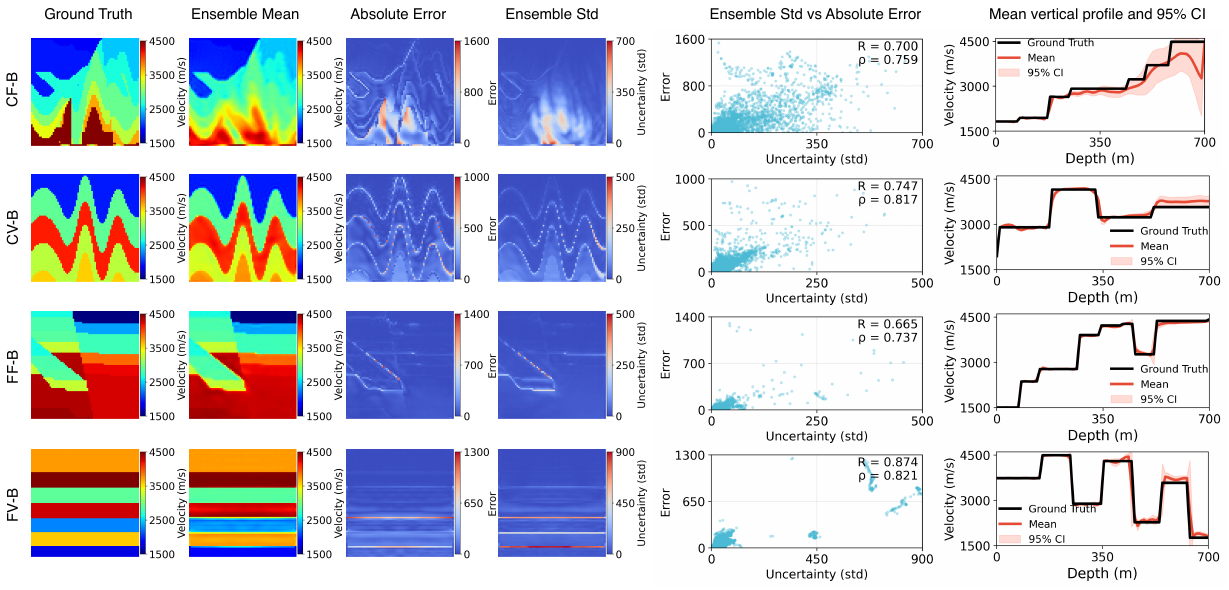}
\caption{\textbf{Examples of uncertainty quantification of RED-DiffEq.} The ensemble mean and standard deviation are computed using an ensemble size of 20. We illustrate the correlation between pixel-wise absolute error and predicted uncertainty, where $R$ represents the Spearman rank, and $\rho$ represents the Pearson correlation. Vertical profiles display the mean prediction with 95\% confidence intervals at the midpoint (350 m) of each velocity model.}
\label{fig:uq_openfwi}
\end{figure}

The results in Fig.~\ref{fig:uq_openfwi} indicate that high uncertainty is concentrated in regions of structural complexity (e.g., fault lines), and this predicted uncertainty exhibits a positive correlation with the absolute reconstruction error measured between the ground truth and the ensemble mean of the reconstructed velocity models. This relationship is further corroborated by the middle panel and vertical profiles in Fig.~\ref{fig:uq_openfwi}. Across all 400 OpenFWI test models, the average Spearman rank ($R$) correlation between absolute error and predicted uncertainty is 0.66, and the average Pearson correlation ($\rho$) is 0.52. Furthermore, the 95\% confidence intervals derived from the ensemble consistently contain the ground-truth profiles for the selected samples, suggesting that the uncertainty estimates are well-calibrated.

\subsection{Validation on the Marmousi and Overthrust benchmarks}
\label{sec:marmousi_result}

To further evaluate our method's performance, we conducted experiments on the widely used Marmousi model~\cite{marmousi}. This challenging benchmark differs substantially from the OpenFWI dataset used to train our diffusion model. While OpenFWI primarily consists of canonical velocity fields with simplified fault geometries and moderate lateral variability, the Marmousi model simulates a deltaic sedimentary environment characterized by rapid sediment accumulation, salt tectonics (salt creep), and growth faults. This provides a test of out-of-distribution performance and domain decomposition capacity on a more complex synthetic dataset with a larger domain.

\begin{figure}[t!]
    \centering
    \includegraphics[width=1\linewidth]{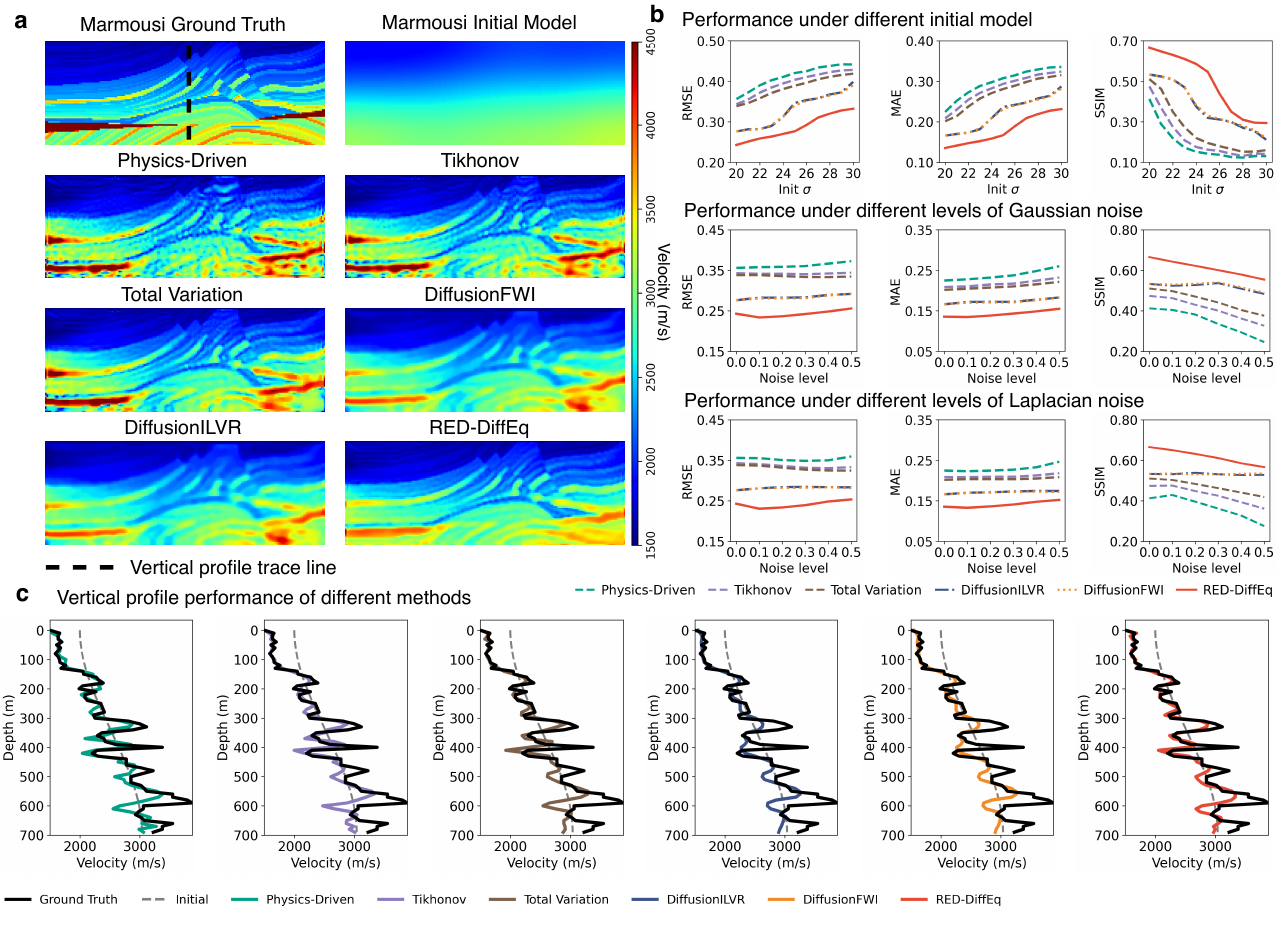}  
    \caption{\textbf{Results on the Marmousi model.} (\textbf{a}) Qualitative comparisons among different methods with initial model $\sigma=20$. (\textbf{b}) Quantitative performance of each method under various initial parameters $\sigma$ (first row), Gaussian noise levels (second row), and Laplacian noise levels (third row). The variable $\sigma$ indicates the standard deviation of the Gaussian filter applied to the ground-truth data to generate the initial model. Larger $\sigma$ implies less preserved information. (\textbf{c}) Vertical velocity profiles at the midpoint position comparing the inversion results to the ground truth.}
    \label{fig:marmousi}
\end{figure}

We initialized the velocity model by applying a Gaussian filter (with $\sigma$ ranging from 20 to 30) to the ground-truth model, effectively removing fine-scale details to mimic practical scenarios characterized by limited prior information. RED-DiffEq accurately reconstructs critical geological features within 300 iterations, recovering complex fault systems, sharp velocity discontinuities, and layered strata (Fig.~\ref{fig:marmousi}a). Crucially, it maintains this high-fidelity performance across all considered initial conditions, as evidenced by the consistently superior results in RMSE, MAE, and SSIM (Fig.~\ref{fig:marmousi}b top row). We note that the baselines DiffusionFWI and DiffusionILVR use the patch-based sliding window approach described in the original paper~\cite{Wang_2023}, which differs from the DDR strategy utilized in RED-DiffEq.

We further evaluated robustness against Gaussian and Laplacian noise in the seismic data where the initial velocity model is obtained by applying a Gaussian filter with $\sigma = 20$. For Gaussian noise with standard deviations ranging from 0.1 to 0.5, the corresponding SNR decreases from 18.42 dB to 4.44 dB. For Laplacian noise with scales ranging from 0.1 to 0.5, the corresponding SNR decreases from 15.41 dB to 1.43 dB. While all methods exhibit reasonable stability in terms of RMSE and MAE (Fig.~\ref{fig:marmousi}b, second and third rows), their performance degrades dramatically in terms of SSIM. Notably, while DiffusionFWI and DiffusionILVR exhibit strong robustness against noise, RED-DiffEq still outperforms all methods across noise levels from 0.1 to 0.5.

To examine velocity reconstructions in greater detail, we compare vertical profiles at midpoint for Marmousi, where results are derived from an initial velocity model generated by applying a Gaussian filter with $\sigma = 20$ (Fig.~\ref{fig:marmousi}c). RED-DiffEq closely follows the ground truth across diverse depth intervals, capturing both sharp velocity transitions and more gradual gradients. Other methods deviate substantially at mid-range to deeper sections, underscoring their difficulties in preserving critical geological features under limited initial information.

We extended our evaluation to the Overthrust model~\cite{overthrust}, which serves as another challenging benchmark that differs from our training data, further testing our method's generalization and domain decomposition capabilities. Specifically, the Overthrust model simulates a complex onshore geological environment featuring large-scale thrust faults, repeated stratigraphic units, and discontinuous velocity fields that are absent in more canonical and simpler velocity model distributions of OpenFWI.

\begin{figure}[t!]
    \centering
    \includegraphics[width=1\linewidth]{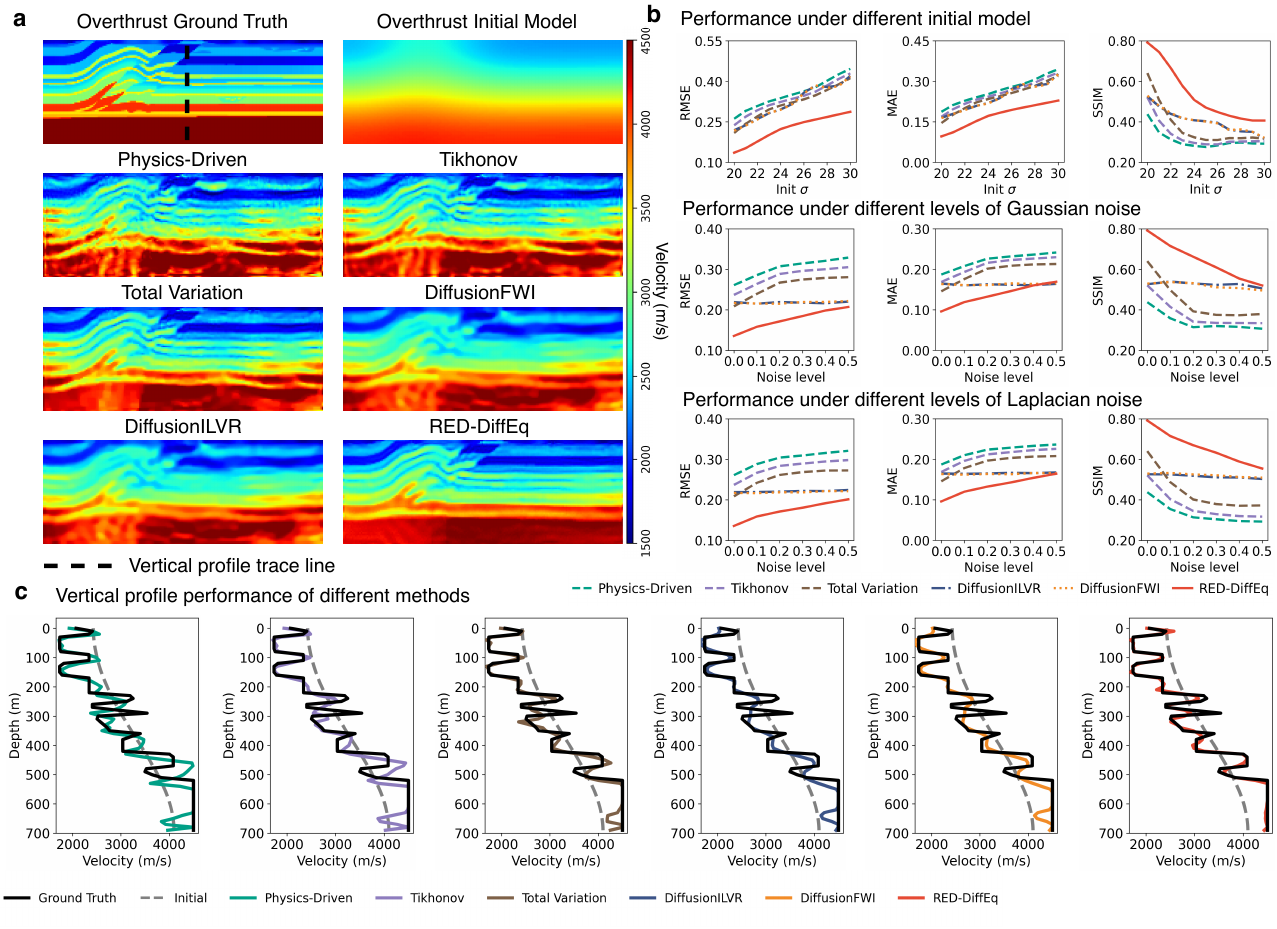}  
    \caption{\textbf{Results on the Overthrust model.} (\textbf{a}) Qualitative comparisons among different methods with initial model $\sigma=20$. (\textbf{b}) Quantitative performance (RMSE, MAE, and SSIM) of each method under various initial parameters $\sigma$ (first row), Gaussian noise levels (second row), and Laplacian noise levels (third row). The variable $\sigma$ indicates the standard deviation of the Gaussian filter applied to the ground-truth data to generate the initial model. Larger $\sigma$ implies less preserved information. (\textbf{c}) Vertical velocity profiles at the midpoint position comparing the inversion results to the ground truth.}
    \label{fig:overthrust}
\end{figure}

Similar to the Marmousi experiments, we initialized the velocity model using a Gaussian filter with $\sigma$ ranging from 20 to 30 applied to the ground-truth model. RED-DiffEq successfully reconstructs the essential geological features within 300 iterations (Fig.~\ref{fig:overthrust}a). The quantitative results in Fig.~\ref{fig:overthrust}b confirm that our method consistently outperforms benchmark methods across all initial conditions in terms of RMSE, MAE, and SSIM. While there is some performance degradation at higher $\sigma$ values, RED-DiffEq remains significantly more robust than other methods.

Similarly, we tested the robustness against Gaussian and Laplacian noise for all methods where results are obtained with an initial velocity model by applying a Gaussian filter with $\sigma = 20$. For Gaussian noise with standard deviations ranging from 0.1 to 0.5, the corresponding SNR decreases from 19.27 dB to 5.28 dB. For Laplacian noise with scales ranging from 0.1 to 0.5, the corresponding SNR decreases from 16.25 dB to 2.28 dB. Unlike the Marmousi case, conventional regularization methods suffer significantly more from noise contamination in this setting, evidenced by worse RMSE, MAE, and SSIM scores. DiffusionFWI and DiffusionILVR continue to exhibit very strong robustness; however, RED-DiffEq still outperforms other methods under all considered scenarios.

The vertical velocity profile at midpoint for the Overthrust model, where results are obtained with an initial velocity model by applying a Gaussian filter with $\sigma = 20$ (Fig.~\ref{fig:overthrust}c), demonstrates RED-DiffEq's ability to accurately capture velocity variations across different depths. Other methods struggle with accurately recovering the velocity structure, particularly in deeper sections. RED-DiffEq's performance on the Overthrust model further confirms its ability to generalize to different geological settings while maintaining high reconstruction fidelity, demonstrating its practical value for real-world seismic imaging applications.

\section{Conclusions}
\label{sec:conclusions}

We introduce RED-DiffEq, a framework that integrates diffusion models as a regularization method for solving inverse PDE problems, with a particular focus on full waveform inversion (FWI). Through comprehensive experiments, we have demonstrated the key advantages of RED-DiffEq. RED-DiffEq not only performs better than other methods on clean seismic data, but also shows remarkable robustness against three types of data imperfections: Gaussian and Laplacian noise contamination and missing traces in the seismic data, where conventional methods typically fail or require careful parameter tuning. In addition, RED-DiffEq shows the capacity for producing meaningful uncertainty quantification. We note, however, that the calibration quality of ensemble-based uncertainty estimates may degrade in out-of-distribution scenarios, where the training data does not fully capture the target geological complexity. RED-DiffEq also demonstrates strong generalization and domain decomposition capabilities beyond its training data distribution and domain size. Despite being trained only on the OpenFWI dataset, it successfully reconstructs complex geological structures with a larger domain size in the challenging Marmousi and Overthrust benchmarks. We note that the experiments in this work are conducted on synthetic benchmarks. Applying RED-DiffEq to real field seismic data requires access to proprietary datasets and careful adaptation to field-specific noise characteristics, acquisition geometries, and significantly larger spatial scales. Addressing these challenges and validating our approach on real field data constitute important directions for future work.

\section{Methods}
\label{sec:methods}

\subsection{RED-DiffEq framework}
\label{sec:red-diffeq}

We present RED-DiffEq as a two-stage framework that integrates a learned prior into a PDE-governed inverse problem via regularization by denoising (RED). The method comprises two stages: (1) a pretraining stage that learns a generative prior and provides an approximation to the marginal score of the noisy data distribution (Fig.~\ref{fig:diffusion} and Sec.~\ref{sec:diffusion_pretraining}), and (2) an inversion stage that uses a Tweedie-inspired denoise operator as a regularizer (Fig.~\ref{fig:diff_fwi_framework}). In the RED regularizer (Eq.~\eqref{eq:red_regularizer}), the operator $\mathcal{D}_\theta$ encourages the current solution $\mathbf{x}_k$ toward regions of higher plausibility under the prior distribution with density $p_{\text{data}}$. 

\paragraph{Stage 1: DDPM pretraining.}
We train a denoising diffusion probabilistic model to predict the Variance-Preserving (VP) corruption noise, which yields an estimator of the marginal score of the noisy prior under the standard $\epsilon$-prediction parameterization (Fig.~\ref{fig:diffusion}a). The VP forward corruption is
\begin{equation} 
\label{eq:method_forward} 
\mathbf{x}_{\text{data},t} \;=\; \sqrt{\gamma(t)}\,\mathbf{x}_{\text{data}} \;+\; \sqrt{1-\gamma(t)}\,\boldsymbol{\epsilon}, \quad \boldsymbol{\epsilon}\sim\mathcal{N}(\mathbf{0},\mathbf{I}),\quad \gamma(t)\in(0,1],
\end{equation} 
which induces the conditional density
\begin{equation}q\!\big(\mathbf{x}_{\text{data},t}\mid\mathbf{x}_{\text{data}}\big) \;=\;\mathcal{N}\!\Big(\mathbf{x}_{\text{data},t};\ \sqrt{\gamma(t)}\,\mathbf{x}_{\text{data}},\ (1-\gamma(t))\,\mathbf{I}\Big), 
\end{equation}
and the noisy marginal 
\begin{equation}
\label{eq:noisy_marginal_integral}
p_{\text{data},t}\!\big(\mathbf{x}_{\text{data},t}\big) =\!\int p_{\text{data}}\!\big(\mathbf{x}_{\text{data}}\big)\, q\!\big(\mathbf{x}_{\text{data},t}\mid\mathbf{x}_{\text{data}}\big)\,d\mathbf{x}_{\text{data}}.
\end{equation}
We train a U-Net $\hat{\boldsymbol{\epsilon}}_\theta(\cdot,t)$ with the standard $\epsilon$-prediction loss: 

\begin{equation}
\mathcal{L}_{\text{DDPM}}(\theta) \;=\; \mathbb{E}_{\,t\sim\mathcal{U}\{1:T\},\,\mathbf{x}_{\text{data}}\sim P_{\text{data}},\,\boldsymbol{\epsilon}\sim\mathcal{N}(\mathbf{0},\mathbf{I})} \!\left[ \left\|\, \boldsymbol{\epsilon} -\hat{\boldsymbol{\epsilon}}_\theta\!\big(\mathbf{x}_{\text{data},t},\,t\big) \,\right\|_2^2 \right].
\end{equation}
Under this parameterization, the score of the noisy marginal (Eq.~\eqref{eq:noisy_marginal_integral}) given an arbitrary noisy sample $\mathbf{z}$ is approximated by
\begin{equation} 
\label{eq:method_score} s_{\text{data},t}\!\big(\mathbf{z}\big) \;=\; \nabla_{\mathbf{z}}\log p_{\text{data},t}\!\big(\mathbf{z}\big) \;\approx\; -\,\frac{\hat{\boldsymbol{\epsilon}}_\theta\!\big(\mathbf{z},t\big)}{\sqrt{1-\gamma(t)}},
\end{equation}
and we provide more details in Sec.~\ref{sec:score_eps}.

\paragraph{Stage 2: Tweedie-inspired denoise operator and inversion.}
During inversion, the prior network is frozen and we use Tweedie’s identity for the Gaussian (VP) channel to define the denoise operator. For a current solution $\mathbf{x}_k$, we sample $t\!\sim\!\mathcal{U}\{1:T\}$ and $\boldsymbol{\epsilon}\!\sim\!\mathcal{N}(\mathbf{0},\mathbf{I})$, compute $\mathbf{x}_{k,t}$ with $\gamma(t)\in(0,1]$ using Eq.~\eqref{eq:vp_noise}, and apply Tweedie’s identity (Eq.~\eqref{eq:tweedie_identity}) to obtain (see Sec.~\ref{sec:VP_Tweedie} for more details) 
\begin{equation} 
\label{eq:method_dtheta} 
\mathcal{D}_\theta(\mathbf{x}_k,t) \;=\; \frac{1}{\sqrt{\gamma(t)}} \Bigl(\mathbf{x}_{k,t} + \bigl(1-\gamma(t)\bigr)\,s_{\text{data},t}(\mathbf{x}_{k,t})\Bigr). 
\end{equation} 
Using the $\epsilon$-prediction surrogate from Eq.~\eqref{eq:method_score}, we evaluate Eq.~\eqref{eq:method_dtheta} in practice as 
\begin{equation} 
\label{eq:method_dtheta_practical} 
\mathcal{D}_\theta(\mathbf{x}_k,t) \;=\; \frac{1}{\sqrt{\gamma(t)}} \Bigl(\mathbf{x}_{k,t} - \sqrt{1-\gamma(t)}\,\hat{\boldsymbol{\epsilon}}_\theta(\mathbf{x}_{k,t},t)\Bigr). \end{equation} 
Plugging Eq.~\eqref{eq:method_dtheta_practical} into the RED regularizer (Eq.~\eqref{eq:red_regularizer}) and expanding with Eq.~\eqref{eq:vp_noise} yields (see Sec.~\ref{sec:algebraic_expansion}) 
\begin{equation} 
\label{eq:method_red_estimator} 
R(\mathbf{x}_k) \;=\; \mathbb{E}_{t,\boldsymbol{\epsilon}} \left[w(t)\,\mathbf{x}_k^\top\!\Bigl(\hat{\boldsymbol{\epsilon}}_\theta(\mathbf{x}_{k,t},t)-\boldsymbol{\epsilon}\Bigr)\right], \quad w(t)\;=\;\sqrt{\frac{1-\gamma(t)}{\gamma(t)}}. 
\end{equation}
During experiments, we use a single-sample Monte Carlo estimate of the expectation. Moreover, we empirically find that dropping $w(t)$ yields better convergence and performance (Sec.~\ref{sec:time_weight_ablation}). Therefore, we fix the regularization weight empirically as a constant $\lambda$.

Our final inversion objective at iteration $k$ combines the data fidelity (PDE misfit) with the RED regularization term: 
\begin{equation} 
\label{eq:method_total_loss} \mathcal{L}(\mathbf{x}_k) \;=\; \bigl\|\,\mathbf{u}_{\text{data}} - f_{\mathrm{PDE}}(\mathbf{x}_k)\,\bigr\|_2^2 \;+\; \lambda\;\mathbf{x}_k^\top\!\Bigl(\hat{\boldsymbol{\epsilon}}_\theta(\mathbf{x}_{k,t},t)-\boldsymbol{\epsilon}\Bigr). 
\end{equation} 
We minimize Eq.~\eqref{eq:method_total_loss} with respect to $\mathbf{x}_k$ using gradient descent optimization. To reduce the computational cost, we stop gradients through the denoiser for computing $\nabla_{\mathbf{x}}R(\mathbf{x}_k)$, i.e., we treat $\hat{\boldsymbol{\epsilon}}_\theta$ as a constant with respect to $\mathbf{x}_k$ in the regularizer term, whose performance has been empirically validated in Refs.~\cite{red-diff,RED2}. At each iteration, we resample $(t,\boldsymbol{\epsilon})$, which introduces stochasticity to the optimization. The algorithm for RED-DiffEq is shown in Algorithm~\ref{alg:fwi-diffq}.

%\paragraph{RED-DiffEq algorithm}
\begin{algorithm}[htbp]
\caption{\textbf{RED-DiffEq for full waveform inversion.}}
\label{alg:fwi-diffq}
\begin{algorithmic}[1]
\REQUIRE Seismic data \(\mathbf{\mathbf{u}_{\text{data}}}\), wave equation solver \(f_{\text{PDE}}(\cdot)\), iterations \(K\), noise schedule \(\{\gamma(t)\}_{t=1}^T\), pre-trained diffusion model \(\hat{\boldsymbol{\epsilon}}_\theta\), regularization parameter \(\lambda\), step size \(\eta\), \(T = 1000\)
\STATE Initialize velocity model: \(\mathbf{x}_0\)
\FOR{\(k = 0, \ldots, K-1\)}
   \STATE \(t\sim\mathcal{U}\{1:T\}\) \COMMENT{Sample timestep uniformly (integer)}
   \STATE \(\boldsymbol{\epsilon} \sim \mathcal{N}(0, \mathbf{I})\) \COMMENT{Sample random Gaussian noise}
   \STATE \(\mathbf{x}_{k,t} = \sqrt{\gamma(t)} \mathbf{x}_k + \sqrt{1 - \gamma(t)} \boldsymbol{\epsilon}\) \COMMENT{Perturb velocity model at iteration \(k\)}
   \STATE \(\hat{\boldsymbol{\epsilon}} = \hat{\boldsymbol{\epsilon}}_\theta(\mathbf{x}_{k,t}, t)\) \COMMENT{Predict noise using pretrained diffusion model}
   \STATE 
   \(
   \mathcal{L}(\mathbf{x}) = \|\mathbf{u}_{\text{data}} - f_{\text{PDE}}(\mathbf{x})\|_2^2 + \lambda \mathbf{x}^\top (\text{sg}(\hat{\boldsymbol{\epsilon}}) - \boldsymbol{\epsilon})
   \)
   \COMMENT{Calculate the loss, stop gradient (sg) on $\hat{\epsilon}$}
   \STATE \(\mathbf{x}_{k+1} \leftarrow \mathbf{x}_k - \eta \nabla_{\mathbf{x}}\mathcal{L}(\mathbf{x})|_{\mathbf{x}=\mathbf{x}_k}\) \COMMENT{Update via gradient descent (e.g., Adam)}
\ENDFOR
\RETURN \(\mathbf{x}_{K}\) \COMMENT{Final velocity model}
\end{algorithmic}
\end{algorithm}

The dominant cost of RED-DiffEq remains in the forward solver and differentiation process, as in conventional FWI. The diffusion-based regularization introduces a single U-Net inference per iteration, which is computationally negligible compared to wave-equation solves with a 0.8\% runtime overhead (Sec.~\ref{sec:computational_cost}).

\subsection{Evaluation metrics}

We assess reconstruction quality using three common metrics: the root mean square error (RMSE), the mean absolute error (MAE), and the structural similarity index measure (SSIM). Let $\{\mathbf{x}_i\}_{i=1}^N$ and $\{\hat{\mathbf{x}}_i\}_{i=1}^N$ denote the ground-truth and recovered models across a spatial domain of $N$ points, respectively. The RMSE and MAE are defined by
\begin{equation}
\begin{aligned}
\text{RMSE} &= \sqrt{\frac{1}{N}\,\sum_{i=1}^N 
\bigl(\mathbf{x}_i - \hat{\mathbf{x}}_i\bigr)^2},
\\[0.5em]
\text{MAE}  &= \frac{1}{N}\,\sum_{i=1}^N 
\bigl|\mathbf{x}_i - \hat{\mathbf{x}}_i\bigr|.
\end{aligned}
\end{equation}
Both measure point-wise differences, with RMSE emphasizing larger errors and MAE capturing average deviations.

SSIM~\cite{ssim} compares structural and perceptual similarities between $\mathbf{x}_i$ and $\hat{\mathbf{x}}_i$. Its value ranges from 0 to 1, where higher scores indicate greater visual and structural fidelity. Formally,
\begin{equation}
\text{SSIM}(\mathbf{x}, \hat{\mathbf{x}}) \;=\; 
\frac{\bigl(2 \mu_x \mu_{\hat{x}} + C_1\bigr)\bigl(2 \sigma_{x\hat{x}} + C_2\bigr)}
{\bigl(\mu_x^2 + \mu_{\hat{x}}^2 + C_1\bigr)\bigl(\sigma_x^2 + \sigma_{\hat{x}}^2 + C_2\bigr)},
\end{equation}
where $\mu_x, \mu_{\hat{x}}$ are the local means of $\mathbf{x}$, $\hat{\mathbf{x}}$, $\sigma_x^2$, $\sigma_{\hat{x}}^2$ are their variances, $\sigma_{x\hat{x}}$ is the covariance, and $C_1, C_2$ are small constants for numerical stability. Note that the local statistics are computed over small sliding windows (e.g., $11\times 11$) centered at each pixel location. The resulting local SSIM values are then averaged over the entire image to yield the final score. While RMSE and MAE quantify overall intensity errors, SSIM captures perceptual and structural consistency, providing complementary insights into the quality of the reconstructed velocity models.

\section*{Acknowledgments}
This work was supported by the U.S. Department of Energy Office of Advanced Scientific Computing Research under Grants No.~DE-SC0025593 and DE-SC0025592.

\section*{Author contributions}
All authors contributed to the conceptualization and design of the study. S.S. and M.Z. implemented the code and conducted the experiments. All authors took part in writing and revising the manuscript. Y.L. and L.L. supervised the research effort, providing guidance and critical feedback throughout.

\section*{Competing interests}
The authors declare no competing interests.

\section*{Code availability}
The codes in this study will be publicly available at the GitHub repository \url{https://github.com/lu-group/red-diffeq}.

\bibliographystyle{unsrt}
\bibliography{main_R2}

\appendix
\clearpage
\renewcommand{\thesection}{S\arabic{section}}
\renewcommand{\thefigure}{S\arabic{figure}}
\renewcommand{\thetable}{S\arabic{table}}
\renewcommand{\theequation}{S\arabic{equation}}
\renewcommand{\thealgorithm}{S\arabic{algorithm}}
\setcounter{figure}{0}
\setcounter{table}{0}
\setcounter{equation}{0}

\section{Acoustic full waveform inversion}
\label{sec:appendix_FWI}

Full waveform inversion (FWI) is an advanced seismic imaging technique that reconstructs high-resolution subsurface velocity models by utilizing the full seismic waveform. Unlike conventional methods that depend on simplified assumptions or isolated wave arrivals, FWI leverages the full physics of seismic wave propagation by directly embedding the governing equations---typically acoustic or elastic wave equations---into the inversion framework.

In the acoustic approximation, commonly used in exploration geophysics, wave propagation is governed by the following second-order PDE:
\begin{equation}
\frac{1}{\mathbf{x}^2(\mathbf{r})} \frac{\partial^2 \mathbf{u}(\mathbf{r},t)}{\partial t^2} - \nabla^2 \mathbf{u}(\mathbf{r},t) = q(\mathbf{r},t),
\end{equation}
where \(\mathbf{u}(\mathbf{r},t)\) represents the seismic wavefield, \(\mathbf{x}(\mathbf{r})\) denotes the spatial distribution of the acoustic velocity model, \(q(\mathbf{r},t)\) is the seismic source term (located at points A through E in the acquisition geometry shown in Fig.~\ref{fig:openfwi}), and \(\nabla^2\) is the Laplacian operator. The solution to this wave equation defines the forward modeling operator \(f(\mathbf{x})\), which maps velocity models to synthetic seismic data.

FWI is formulated as a nonlinear optimization problem aimed at determining the velocity model \(\mathbf{x}\) that minimizes the difference between the observed seismic data \(\mathbf{u}_{\text{data}}\) and the synthetic data \(f(\mathbf{x})\) generated by solving the wave equation numerically. This optimization problem can be expressed as
\begin{equation}
\arg\min_{\mathbf{x}} \left[ \frac{1}{2} \| \mathbf{u}_{\text{data}} - f_{\text{PDE}}(\mathbf{x}) \|_2^2 + \lambda R(\mathbf{x}) \right],
\end{equation}
where \(\| \cdot \|_2\) denotes the Euclidean norm, \(R(\mathbf{x})\) represents a regularization term that incorporates prior information or enforces specific model characteristics, and \(\lambda\) is a regularization parameter that balances data fidelity with regularization constraints.

The physics-driven nature of FWI is evident in the forward modeling operator \(f(\mathbf{x})\), which requires solving the acoustic wave equation using numerical methods such as finite differences. In this process, synthetic seismic data is generated by simulating wave propagation from known source positions and recording the resulting wavefield at receiver locations, replicating the actual data acquisition geometry.

Despite its advantages, FWI presents significant computational and practical challenges due to its inherent nonlinearity and ill-posed nature. These challenges make FWI highly sensitive to noise and inaccuracies in the initial model. Noise in seismic data acquisition is inevitable and can severely degrade inversion performance. To mitigate these effects and improve the robustness of the inversion process, different regularization techniques are employed. These techniques integrate prior geological information and enforce model constraints, resulting in physically plausible velocity models.

\clearpage
\section{Noise schedule algorithm}
\label{sec:appendix_algorithm}
\begin{figure}[htbp]
    \centering
    \includegraphics[width=1\linewidth]{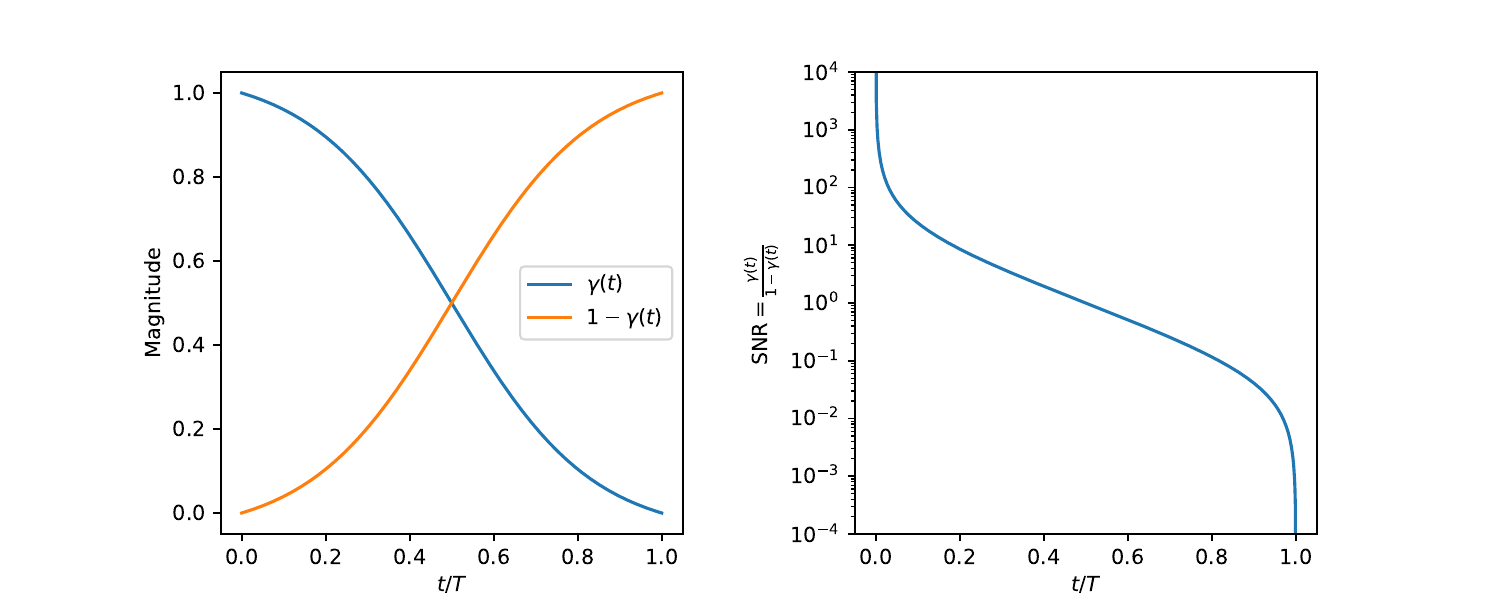}  
    \caption{\textbf{Sigmoid noise schedule.} (\textbf{Left}) Cumulative signal‑retention factor $\gamma(t)$ and its complement $1-\gamma(t)$, illustrating the decay of clean signal versus accumulated noise. (\textbf{Right}) Signal‑to‑noise ratio $\gamma(t)/(1-\gamma(t))$, illustrating the transition from high‑SNR to low‑SNR regimes over time.}
    \label{fig:noise_schedule}
\end{figure}

The noise schedule \(\{\gamma(t)\}_{t=1}^T\) specifies the cumulative signal‐ratio at each diffusion step \(t=1,\dots,T\). In diffusion models, the noise schedule dictates how noise is progressively added to the input data over \(T\) time steps, transforming it into pure noise in the forward process. This schedule is critical for training the model to reverse the process and recover the original signal. In our implementation, we adopt a sigmoid‑based \(\gamma\)–schedule~\cite{sigmoid_noise_schedule} (Algorithm~\ref{alg:sigmoid_noise_schedule} and Fig.~\ref{fig:noise_schedule}), as it demonstrates superior generation performance compared to other noise schedules (e.g., linear and cosine).

\begin{algorithm}[htbp]
\caption{\textbf{Sigmoid \(\gamma\) schedule.}}
\label{alg:sigmoid_noise_schedule}
\begin{algorithmic}[1]
  \REQUIRE Total steps \(T=1000\), start \(S=-3\), end \(E=3\), sharpness \(\tau=1\)
  \STATE \(\sigma(x) = \tfrac{1}{1 + e^{-x}}\)
  \STATE \(v_S \gets \sigma\!\bigl(\tfrac{S}{\tau}\bigr), \quad v_E \gets \sigma\!\bigl(\tfrac{E}{\tau}\bigr)\)
  \FOR{\(t = 1\) to \(T\)}
    \STATE \(u_t \gets \tfrac{t}{T}\)
    \STATE \(s_t \gets \tfrac{u_t\,(E - S) + S}{\tau}\)
    \STATE \(\gamma_t \gets \dfrac{\,v_E - \sigma(s_t)\,}{\,v_E - v_S\,}\)
  \ENDFOR
  \RETURN \(\{\gamma_t\}_{t=1}^T\)
\end{algorithmic}
\end{algorithm}

In diffusion models, the signal-to-noise ratio (SNR) at each step quantifies the relative strength of the remaining clean data versus the accumulated noise during the forward diffusion process. It is defined as \(\mathrm{SNR}(t) = \gamma(t) / (1 - \gamma(t))\), and thus \(\gamma(t)\) is also called the cumulative signal-retention factor. As diffusion progresses, the model learns to predict increasingly noisy data, and the SNR directly influences the training dynamics. The sigmoid noise schedule causes a rapid transition from high-SNR (signal-dominant) to low-SNR (noise-dominant) regimes.

\clearpage
\section{Mathematical framework of RED-DiffEq}
\label{sn:vp_identities}

Throughout, let $\gamma(t)\in(0,1]$ denote the variance-preserving (VP) noise schedule at step $t\in\{1,\dots,T\}$. Let $p_{\text{data}}$ be the density of the clean prior distribution.

\subsection{Score function by DDPM}
\label{sec:score_eps}
The marginal score is defined as 
\begin{equation}
s_{\text{data},t}(\mathbf{z}) \coloneq 
\nabla_{\mathbf{z}}\log p_{\text{data},t}(\mathbf{z}).
\end{equation}
Under the DDPM framework~\cite{ddpm}, for a network $\hat{\boldsymbol{\epsilon}}_\theta(\mathbf{z},t)$ trained with the standard $\epsilon$-prediction objective, we have the approximation
\begin{equation}
\label{eq:score_approx}
s_{\text{data},t}(\mathbf{z}) \;\approx\; 
-\,\frac{\hat{\boldsymbol{\epsilon}}_\theta(\mathbf{z},t)}{\sqrt{1-\gamma(t)}}.
\end{equation}

\subsection{Posterior mean for the VP process}
\label{sec:VP_Tweedie}
Let $\mathbf{y} = \mathbf{x} + \sigma\,\boldsymbol{\epsilon}$ with $\boldsymbol{\epsilon}\!\sim\!\mathcal{N}(\mathbf{0},\mathbf{I})$ and the noise variance $\sigma^2$. Tweedie’s formula~\cite{efron2011tweedie} states
\begin{equation}
\mathbb{E}\!\big[\mathbf{x} \,\big|\, \mathbf{y}\big]
\;=\;\mathbf{y}+ \sigma^2\nabla_{\mathbf{y}}\log p(\mathbf{y}).
\end{equation}
Under the VP channel $\mathbf{x}_t=\sqrt{\gamma(t)}\,\mathbf{x}+\sqrt{1-\gamma(t)}\,\boldsymbol{\epsilon}$, this gives
\begin{equation}
\begin{aligned}
\label{eq:sn_tweedie}
\mathbb{E}\!\big[\mathbf{x} \,\big|\, \mathbf{x}_t\big]
\;&=\;
\frac{1}{\sqrt{\gamma(t)}}\Big[\mathbf{x}_t + \big(1-\gamma(t)\big)\,\nabla_{\mathbf{x}_t}\log p_{\text{data},t}(\mathbf{x}_t)\Big] \\
\;&=\;
\frac{1}{\sqrt{\gamma(t)}}\Big[\mathbf{x}_t + \big(1-\gamma(t)\big)\,s_{\text{data},t}(\mathbf{x}_t)\Big].
\end{aligned}
\end{equation}

\subsection{RED estimator}
\label{sec:algebraic_expansion}

Let $\mathbf{x}\in\mathbb{R}^n$ be the current solution during inversion. Using Eqs.~\eqref{eq:score_approx} and \eqref{eq:sn_tweedie}, a practical denoiser is
\begin{equation}
\mathcal{D}_\theta(\mathbf{x};t,\boldsymbol{\epsilon})
\;=\;
\frac{1}{\sqrt{\gamma(t)}}\Big(\mathbf{x}_t \;-\; \sqrt{1-\gamma(t)}\,\hat{\boldsymbol{\epsilon}}_\theta(\mathbf{x}_t,t)\Big),
\quad
\mathbf{x}_t=\sqrt{\gamma(t)}\,\mathbf{x}+\sqrt{1-\gamma(t)}\,\boldsymbol{\epsilon}.
\end{equation}
Then for the regularization $R(\mathbf{x}) = \mathbb{E}_{t,\boldsymbol{\epsilon}}\big[\mathbf{x}^\top\!\bigl(\mathbf{x}-\mathcal{D}_\theta(\mathbf{x};t,\boldsymbol{\epsilon})\bigr)\big]$, we estimate it with one Monte Carlo sample to get

\begin{equation}
\begin{aligned}
\mathbf{x}^{\!\top}\!\big(\mathbf{x}-\mathcal{D}_\theta(\mathbf{x};t,\boldsymbol{\epsilon})\big)
&= \mathbf{x}^{\!\top}\!\left[\mathbf{x}-\frac{\sqrt{\gamma(t)}\,\mathbf{x}+\sqrt{1-\gamma(t)}\,\boldsymbol{\epsilon}-\sqrt{1-\gamma(t)}\,\hat{\boldsymbol{\epsilon}}_\theta(\mathbf{x}_t,t)}{\sqrt{\gamma(t)}}\right] \nonumber\\
&= \sqrt{\frac{1-\gamma(t)}{\gamma(t)}}\;\mathbf{x}^{\!\top}\!\Big(\hat{\boldsymbol{\epsilon}}_\theta(\mathbf{x}_t,t)-\boldsymbol{\epsilon}\Big) \\
&\coloneq
w(t)\,\mathbf{x}^{\!\top}\!\Big(\hat{\boldsymbol{\epsilon}}_\theta(\mathbf{x}_t,t)-\boldsymbol{\epsilon}\Big).
\end{aligned}
\end{equation}

\clearpage

\section{Experiment details and baseline methods}
\label{sec:appendix_experiment_detail}
\subsection{Diffusion model pre-training}
\label{sec:diffusion_pretraining}
The diffusion model was trained on the OpenFWI dataset~\cite{openfwi}, encompassing FF-B, FV-B, CV-B, and CF-B velocity model families, using a U-Net architecture with ResBlock and Attention Block components (Fig.~\ref{fig:diffusion}a). The model architecture incorporated 1,000 diffusion steps with a sigmoidal noise schedule (start = $-3$, end = 3, $\tau=1$) in Alg.~\ref{alg:sigmoid_noise_schedule}, 64 base channels with progressive channel multipliers (1, 2, 4, 8), and 4 attention heads. We trained the model for 600,000 iterations (parameter update steps over mini-batches) using a batch size of 32 and a learning rate of 0.0002.

\begin{figure}[htbp]
    \centering
    \includegraphics[width=0.8\linewidth]{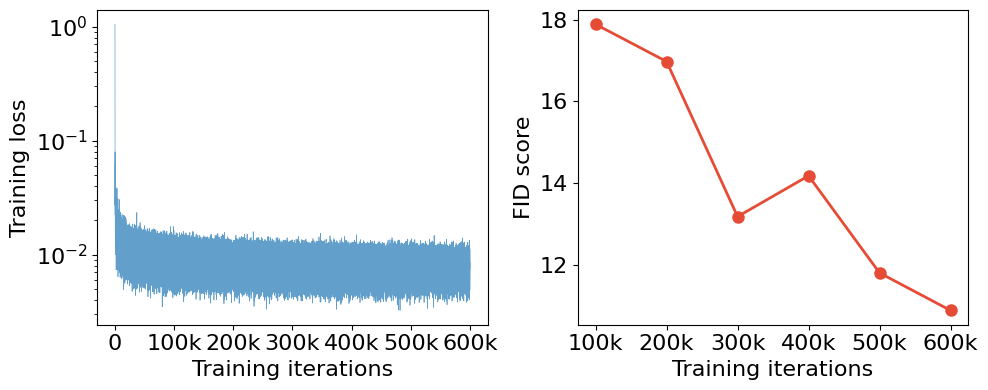}
    \caption{\textbf{Training dynamics of the diffusion model.} (\textbf{Left}) Training loss of the diffusion model. (\textbf{Right}) Fr\'echet Inception Distance (FID) evaluated during training.}
    \label{fig:diffusion_training}
\end{figure}

\begin{figure}[htbp]
    \centering
    \includegraphics[width=1.0\linewidth]{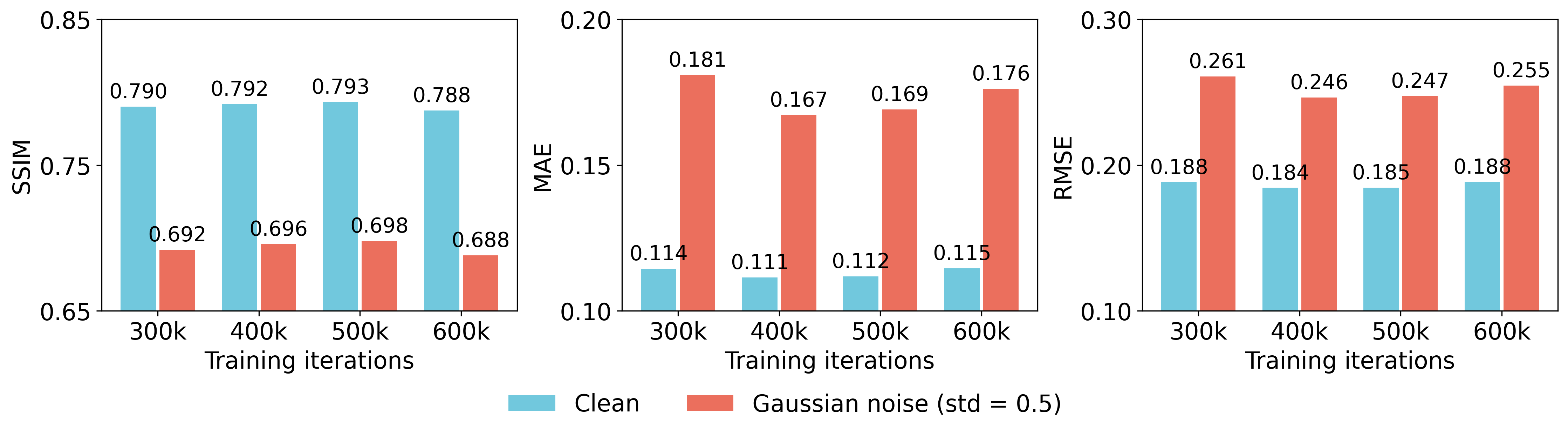}
    \caption{\textbf{Inversion performance of diffusion models trained with different iterations.}}
    \label{fig:checkpoints_performance}
\end{figure}

The training loss converged after approximately 300,000 iterations and continued to oscillate (Fig.~\ref{fig:diffusion_training}, left), which is typical behavior for diffusion model training. We evaluated the Fr\'echet Inception Distance (FID) score to assess generation quality every 100,000 iterations (Fig.~\ref{fig:diffusion_training}, right). While the FID score generally decreases as training progresses, we found that optimal inversion performance does not necessarily align with the lowest FID. Specifically, inversion accuracy is the highest at around 400,000 to 500,000 iterations for both clean seismic data and data contaminated with Gaussian noise ($\sigma=0.5$) (Fig.~\ref{fig:checkpoints_performance}). Therefore, all the diffusion-model-based results reported in this paper are produced using the network trained with 400,000 iterations.

\subsection{Hyper-parameters tuning and design choices}
\label{sec:hyperparameters_tuning}

To ensure fair comparison, hyperparameters were kept consistent across all experiments. The optimization process used a learning rate of 0.03 with a cosine annealing scheduler for 300 iterations, we adopt the Adam optimizer for implementing gradient descent on the velocity model, since in RED-DiffEq the optimization objective incorporates a diffusion-based regularization term that introduces stochasticity through random sampling during each iteration, which makes methods like L-BFGS a less optimal choice. Regularization coefficients were tuned through empirical validation: 0.01 for both Tikhonov and Total Variation methods, and 0.75 for RED-DiffEq. These values were selected to achieve a balance between quantitative accuracy and qualitative fidelity, ensuring the regularization terms effectively constrained the solution space while avoiding common artifacts (such as the staircase effect) that compromise geological interpretation.

For the baseline methods DiffusionFWI and DiffusionILVR, we utilized the official implementations from their respective repositories, adapting the source code to support acoustic FWI (replacing the original elastic engine) and to track intermediate inversion results. Hyperparameters were separately fine-tuned on subsets of the OpenFWI and Marmousi datasets. While the authors of DiffusionFWI and DiffusionILVR proposed several stabilization techniques, including gradient smoothing, gradient normalization, and velocity model blurring, our ablation studies indicated that combining gradient smoothing with velocity model blurring yielded the best results in our setting. Consequently, all reported results for DiffusionFWI and DiffusionILVR incorporate these two techniques.

\subsection{Time dependent weight ablation study}
\label{sec:time_weight_ablation}

We empirically found that replacing the time-dependent weighting $w(t)$ with a fixed parameter $\lambda$ improves performance. Specifically, using a fixed regularization strength yielded an SSIM of 0.7920, MAE of 0.1114, and RMSE of 0.1845. In contrast, the time-weighted approach resulted in an SSIM of 0.7677, MAE of 0.1275, and RMSE of 0.1979. Consequently, the fixed-weight strategy achieves a relative improvement of approximately 3.06\% in SSIM, 14.43\% in MAE, and 7.27\% in RMSE compared to the time-weighted baseline.

\subsection{Forward solver details}
\label{sec:forward_solver}

The 2D acoustic wave equation is solved using a 4th-order finite difference scheme with a Ricker wavelet source. Sources and receivers are uniformly distributed along the surface. The solver configuration is identical to the one used to generate the OpenFWI dataset (Table~\ref{tab:forward_solver_config}).

\begin{table}[htbp]
\centering
\caption{\textbf{Forward solver configuration.}}
\label{tab:forward_solver_config}

\small
\begin{tabular}{@{}llccc@{}}
\toprule
\textbf{Parameter} & \textbf{Symbol} & \textbf{OpenFWI} & \textbf{Marmousi/Overthrust} & \textbf{Unit} \\
\midrule
Horizontal grid points      & $n_x$           & 70   & 190  & -- \\
Vertical grid points        & $n_z$           & 70   & 70   & -- \\
Grid spacing                & $\Delta x$      & 10.0 & 10.0 & m \\
Time steps                  & $n_t$           & 1000 & 1000 & -- \\
Time step size              & $\Delta t$      & 0.001& 0.001& s \\
Recording time              & $T$             & 1.0  & 1.0  & s \\
Source frequency (Ricker)   & $f$             & 15.0 & 15.0 & Hz \\
Number of sources           & $n_s$           & 5    & 5    & -- \\
Number of receivers         & $n_g$           & 70   & 190  & -- \\
Source depth                & $z_s$           & 10   & 10   & m \\
Receiver depth              & $z_g$           & 10   & 10   & m \\
Boundary condition          & -- & \multicolumn{2}{c}{Sponge layer (quadratic damping)} & -- \\
ABC layer thickness         & $n_{\text{bc}}$ & 120  & 120 & grid points \\
\midrule
Physical domain size        & -- & $700\times700$   & $700\times1900$ & m$^2$ \\
Seismic data shape          & -- & $(5,1000,70)$  & $(5,1000,190)$ & -- \\
\bottomrule
\end{tabular}

\end{table}

\subsection{Computational cost for RED-DiffEq}
\label{sec:computational_cost}
The runtime of RED-DiffEq is summarized in Table~\ref{tab:red_diffeq_timing}. RED-DiffEq introduces only a negligible overhead compared to standard FWI without regularization. The additional diffusion step contributes a minor cost, resulting in an overall runtime increase of just 0.8\% per iteration. This indicates that incorporating RED-DiffEq has minimal impact on wall-clock time.

In terms of memory usage, RED-DiffEq requires moderately higher GPU memory due to the storage of diffusion model parameters. The peak VRAM usage increases from 5128~MB to 5884~MB, corresponding to a 14.7\% overhead. However, this increase remains well within the capacity of modern GPUs and does not represent a practical limitation. Overall, RED-DiffEq achieves its regularization benefits with minimal additional computational and memory costs.

\begin{table}[htbp]
\centering
\caption{\textbf{Average runtime breakdown for RED-DiffEq and standard FWI without regularization on the Overthrust dataset (70$\times$190 grid).}
Measured on an NVIDIA RTX 3090 (24GB) and averaged over 500 iterations.
RED-DiffEq introduces only a 0.8\% runtime overhead compared to unregularized FWI.}
\label{tab:red_diffeq_timing}
\small
\begin{tabular}{llc}
\toprule
& Process & Avg. Time (s) \\
\midrule

No Regularization
& Forward solver   & 0.7651 \\
& Differentiation  & 2.0306 \\
& \textbf{Total} & \textbf{2.7958} \\

\midrule

RED-DiffEq
& Forward solver   & 0.7643 \\
& Diffusion step   & 0.0162 \\
& Differentiation  & 2.0368 \\
& \textbf{Total} & \textbf{2.8174} \\

\midrule
\multicolumn{2}{l}{Overhead} & +0.8\% \\

\bottomrule
\end{tabular}
\end{table}

\subsection{Tikhonov regularization} 
\label{sec:tikhonov_regularization}
Tikhonov regularization~\cite{tv_l2} is a widely used technique that promotes smoothness in the velocity model by penalizing large variations in parameter values. We apply the first-order Tikhonov regularization as
\[
R_{\text{Tikhonov}}(\mathbf{x}) = \frac{1}{N} \sum_{i,j} \left((\mathbf{x}_{i+1,j} - \mathbf{x}_{i,j})^2+ (\mathbf{x}_{i,j+1} - \mathbf{x}_{i,j})^2 \right),
\]
where \(\mathbf{x}_{i,j}\) represents the velocity value at the discrete position \((i,j)\), and \(N\) is the total number of grid points. \(R_{\text{Tikhonov}}(\mathbf{x})\) discourages abrupt changes in the velocity field, resulting in a smoother solution. This method is advantageous when the true subsurface structures are expected to vary smoothly.

\subsection{Total variation regularization} 
\label{sec:tv_regularization}
Total Variation (TV) 
regularization~\cite{total_variation} is a powerful technique for preserving sharp interfaces and discontinuities in the velocity model, which are critical for representing geological boundaries such as faults. In our implementation, we apply the anisotropic TV regularization as
\[
R_{\text{TV}}(\mathbf{x}) = \frac{1}{N} \sum_{i,j} \left( |\mathbf{x}_{i+1,j} - \mathbf{x}_{i,j}| + |\mathbf{x}_{i,j+1} - \mathbf{x}_{i,j}| \right),
\]
where \(\mathbf{x}_{i,j}\) represents the velocity value at the discrete position \((i,j)\), and \(N\) is the total number of grid points. However, a common drawback of TV regularization is the potential introduction of staircase artifacts, where smooth gradients are approximated by discrete, flat regions separated by abrupt transitions.

\clearpage
\section{Effect of the post-processing refinement step}
\label{sec:post_process}

\begin{figure}[htbp]
    \centering
    \includegraphics[width=1.0\linewidth]{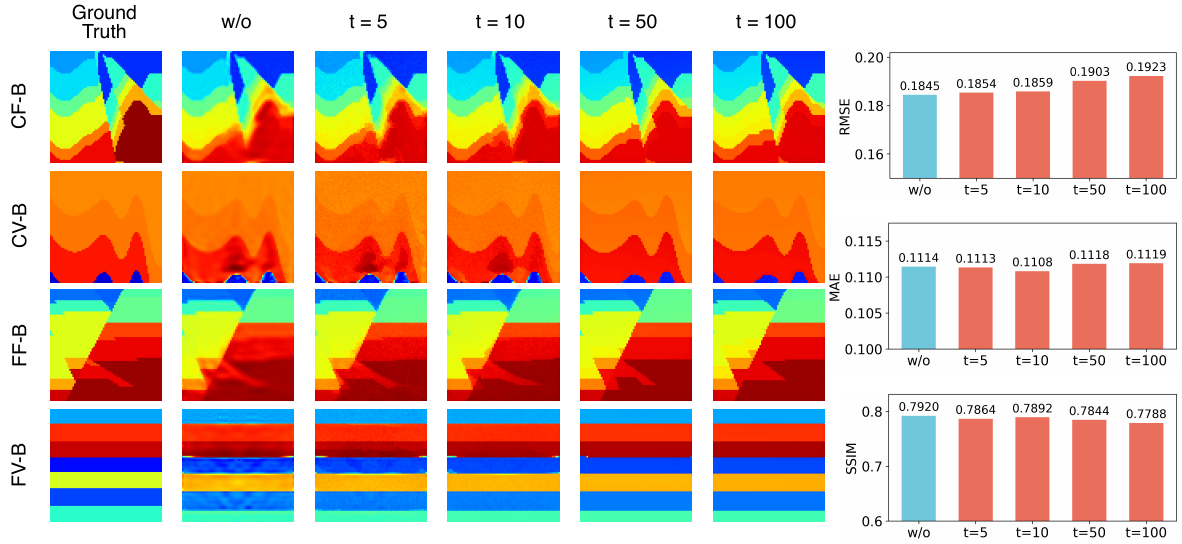}
    \caption{\textbf{The effect of the post-processing step across different diffusion timesteps ($t$).} ``w/o'' denotes the result from RED-DiffEq without any post-processing. The variable $t$ indicates the starting diffusion timestep selected for denoising.}
    \label{fig:post_process}
\end{figure}

We evaluate a diffusion-based post-processing refinement step that denoises the reconstructed velocity model using the pretrained diffusion prior, notably without enforcing data fidelity constraints. Starting from the final inverted model, we apply a forward diffusion process up to a chosen timestep $t$, followed by the conventional DDPM reverse process to obtain a refined estimate.  While increasing $t$ generally strengthens the prior influence and yields visually cleaner and more geologically plausible structures (Fig.~\ref{fig:post_process}, left), the absence of a data-consistency term implies that an overly large $t$ can bias the solution away from the ground truth. This trade-off is reflected in the modest degradation of quantitative metrics (RMSE, MAE, SSIM) observed at higher $t$ values (Fig.~\ref{fig:post_process}, right). Because this refinement step does not enforce data fidelity, it inherently drives the velocity model toward high-probability structures under the learned prior. This strong prior influence can introduce biased features, thereby increasing quantitative errors.

\end{document}